\documentclass[10pt,twocolumn,letterpaper]{article}

\usepackage{multirow}
\usepackage[pagenumbers]{cvpr} 

\usepackage{pifont} 
\usepackage{amsmath}
\usepackage{amssymb}
\usepackage{subcaption}
\usepackage{flushend}
\newcommand{\Yes}{\textcolor{Green}{\ding{51}}}
\newcommand{\No}{\textcolor{Maroon}{\ding{55}}}

\definecolor{cvprblue}{rgb}{0.21,0.49,0.74}
\usepackage[pagebackref,breaklinks,colorlinks,allcolors=cvprblue]{hyperref}

\def\paperID{} 
\def\confName{CVPR}
\def\confYear{2026}

\title{VidTAG: Temporally Aligned Video to GPS Geolocalization with Denoising Sequence Prediction at a Global Scale}

\author{Parth Parag Kulkarni\textsuperscript{1}, Rohit Gupta\textsuperscript{1,3}, Prakash Chandra Chhipa\textsuperscript{2}, Mubarak Shah\textsuperscript{1} \\
\textsuperscript{1}Institute of Artificial Intelligence, University of Central Florida, USA\\
\textsuperscript{2}Luleå Tekniska Universitet, Sweden, \textsuperscript{3}Amazon Prime Video Science, USA
\\
{\tt\small \{parthparag.kulkarni, rohit.gupta\}@ucf.edu, prakash.chandra.chhipa@ltu.se, shah@crcv.ucf.edu}
}

\begin{document}
\maketitle
\begin{abstract}
The task of video geolocalization aims to determine the precise GPS coordinates of a video’s origin and map its trajectory; with applications in forensics, social media, and exploration. Existing classification-based approaches operate at a coarse city-level granularity and fail to capture fine-grained details, while image retrieval methods are impractical on a global scale due to the need for extensive image galleries which are infeasible to compile. Comparatively, constructing a gallery of GPS coordinates is straightforward and inexpensive. We propose VidTAG, a dual-encoder framework that performs frame-to-GPS retrieval using both self-supervised and language-aligned features. To address temporal inconsistencies in video predictions, we introduce the TempGeo module, which aligns frame embeddings, and the GeoRefiner module, an encoder-decoder architecture that refines GPS features using the aligned frame embeddings. Evaluations on Mapillary (MSLS) and GAMa datasets demonstrate our model’s ability to generate temporally consistent trajectories and outperform baselines, achieving a 20\% improvement at the 1 km threshold over GeoCLIP. We also beat current State-of-the-Art by 25\% on global coarse grained video geolocalization (CityGuessr68k). Our approach enables fine-grained video geolocalization and lays a strong foundation for future research. More details on  the project webpage: \href{https://parthpk.github.io/vidtag_webpage/}{parthpk.github.io/vidtag\_webpage}.
\end{abstract}    
\vspace{-1em}
\section{Introduction}
\label{sec:intro}
Geolocalization refers to the procedure of ascertaining the geographic position using visual input. Given an image/video, determining its location in the world can be a valuable asset. Geolocalization has a lot of downstream applications in forensics, like investigative analysis of pictures; social media, like location tagging; and exploration, like study of unknown and dangerous places/landscapes. Current methods fall into two regimes with complementary strengths. Fine‑grained approaches (city‑ or region‑level) predominantly use retrieval: a query image is matched to a geo‑referenced image gallery, and the query inherits the location of its best match. This strategy yields high accuracy but requires heavy compute and large reference galleries, and it is sensitive to domain shift \cite{zhu2022transgeo,mi2024congeo,deuser2023sample4geo,yang2021cross,zhu2021vigor}. Worldwide approaches typically use classification: the Earth is partitioned into regions, and the model predicts a region label. This formulation enables single‑pass inference but provides only approximate locations; increasing granularity raises computational cost and class confusion~\cite{pramanick2022world,clark2023we,haas2024pigeon,muller2018geolocation,vivanco2024geoclip,kulkarni2024cityguessr}. Bridging these extremes, GeoCLIP\cite{vivanco2024geoclip} embeds images and GPS coordinates in a shared space, enabling direct retrieval of GPS points without massive image galleries and making large GPS galleries practical.

\begin{figure}[t]
  \centering
  \includegraphics[width =0.48\textwidth]{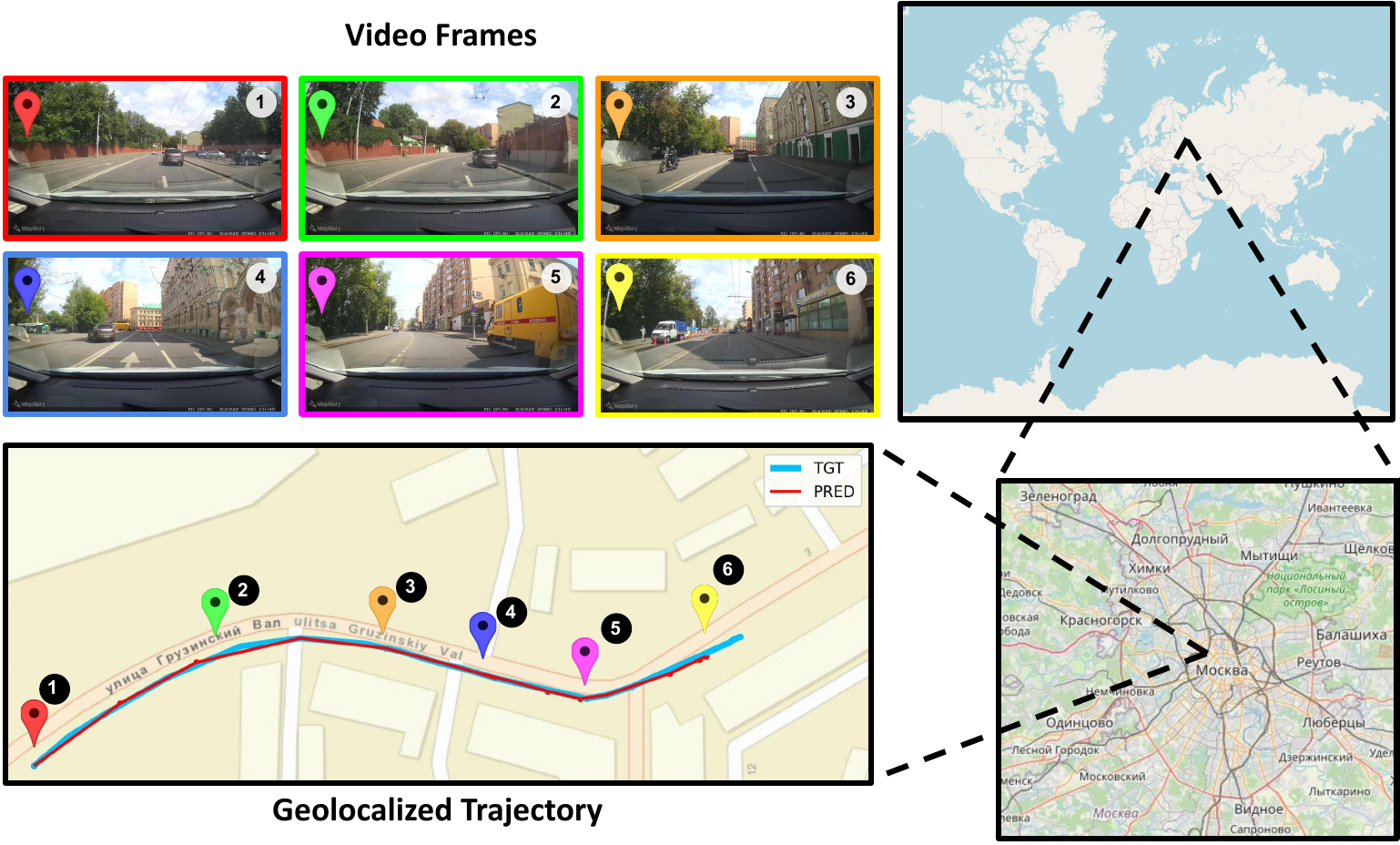}
  \caption{\textbf{Frame-wise Video to GPS Sequence Prediction.} Our model geolocalizes every frame in a video on a global scale to give a temporally consistent trajectory}
  \label{fig:teaser}
  \vspace{-1em}
\end{figure}

\begin{figure}[t]
  \centering
  \includegraphics[width =0.48\textwidth]{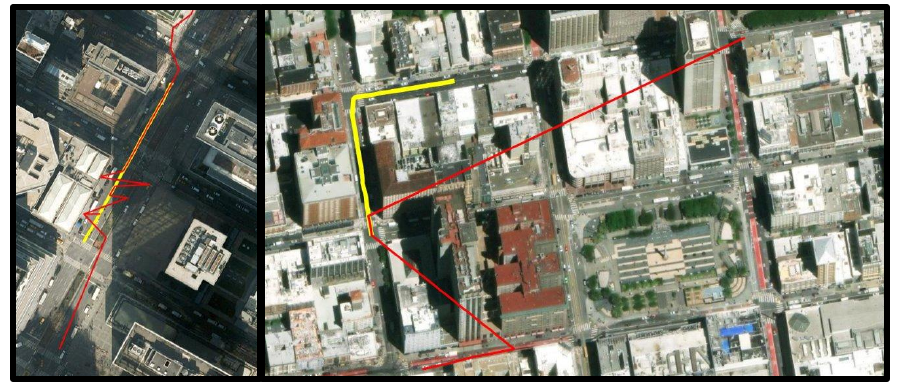}
  \vspace{-2em}
  \caption{\textbf{Temporal Inconsistency Issue.} Geolocalizing each frame separately leads to incoherent trajectories. The problem becomes worse as the scale increases. Yellow: GT, Red: Predicted.}
    \vspace{-1.5em}
  \label{fig:tempinc}
\end{figure}

This inspires us to introduce a novel formulation of this problem of video geolocalization via a frame to GPS retrieval approach. For videos, the central challenge is maintaining temporal consistency in predictions. An ideal model should predict the location of every frame in a way which results in a coherent trajectory. Fig. \ref{fig:teaser} shows a predicted trajectory which aligns with the ground truth trajectory and the model understands the temporal progression of the video frames. Applying image methods frame by frame often produces jittery trajectories(Fig. \ref{fig:tempinc}), and in the worst cases the predicted path jumps across continents. The only global scale video effort to date CityGuessr\cite{kulkarni2024cityguessr}, reasons at the whole‑video level and therefore does not enforce frame‑wise consistency. Consequently, it remains unclear how to obtain precise and temporally coherent trajectories at global scale. 

In this paper, we propose a frame-to-GPS retrieval approach for video geolocalization at a large scale that outputs a temporally consistent trajectory. Our proposed method comprises of a dual-encoder that leverages the different capabilities of DINOv2\cite{oquab2023dinov2} and CLIP\cite{radford2021learning}, along with a TempGeo module for temporal alignment of features of frames of a video, as well as an encoder-decoder module named GeoRefiner, for refining GPS features in accordance with the aligned frame features. As this is the first attempt to solve framewise video geolocalization with a GPS retrieval approach to the best of our knowledge, it can serve as a baseline approach for future research.

\noindent Our main contributions can be highlighted as follows:
\begin{itemize}
    \item We introduce a novel formulation of framewise video geolocalization on a global scale
    \item We propose the first frame-to-GPS retrieval method to solve large-scale frame-wise video geolocalization
    \item Our model comprises of
    \begin{itemize}
        \item dual CLIP and DINOv2 frame encoders combining semantic and visual features
        \item TempGeo and GeoRefiner modules which specifically address the issue of temporal inconsistency
    \end{itemize}
    \item We outperform all state-of-the-art models and baselines on MSLS\cite{warburg2020mapillary} by 20\% @1km, as well as other video datasets such as GAMa\cite{vyas2022gama} by 25\% @1km and CityGuessr68k\cite{kulkarni2024cityguessr} by 25\% at city-level.
\end{itemize}
\section{Related Work}
\label{sec:related}
Video geolocalization has always been primarily retrieval-based and confined to fine-grained localization. Worldwide geolocalization being mostly classification-based, has been relatively unexplored for videos, especially for frame-based geolocalization and trajectory mapping.

\subsection{Image Geolocalization}
\label{sec:related-wig}

\noindent \textbf{Global, primarily classification-based methods.}
Weyand et al.\ \cite{weyand2016planet} introduced worldwide image geolocalization as a classification problem on Im2GPS \cite{hays2008im2gps}. Vo et al.\ \cite{vo2017revisiting} extended this formulation with multi-level geographic hierarchies, while CPlaNet \cite{seo2018cplanet} combined coarse partitions via a combinatorial scheme to predict finer cells. Subsequent work improved the visual encoders and training pipelines: ISNs \cite{muller2018geolocation} encoded scene types explicitly and proposed hierarchical evaluation; Translocator \cite{pramanick2022world} adopted twin encoders for images and segmentations; GeoDecoder \cite{clark2023we} introduced a query-based encoder–decoder; and PIGEON \cite{haas2024pigeon} leveraged the CLIP \cite{radford2021learning} vision encoder with a clustering strategy. GeoCLIP \cite{vivanco2024geoclip} reframed geolocalization as contrastive multimodal learning between images and raw GPS, yielding a strong global-scale \emph{retrieval} model. Following that, GT-Loc\cite{shatwell2025gt} improved upon this strategy by incorporating time prediction. Recently, multi-modal large language models (MLLMs) like Qwen2.5-VL\cite{bai2025qwen2} have shown promise in this domain. Some MLLMs like GeoReasoner\cite{li2024georeasoner} and GAEA\cite{campos2025gaea} are specifically trained for this task. New works like G3\cite{jia2024g3} and GeoRanker\cite{jia2025georanker} have incorporated RAG\cite{lewis2020retrieval} to the MLLM architecture towards improved performance. These methods operate on individual images and do not address the temporal inconsistency intrinsic to video (Sec.~\ref{sec:intro}). Most of the above were trained on MP-16 \cite{larson2017benchmarking}, except CPlaNet \cite{seo2018cplanet}.

\noindent \textbf{Fine-grained, retrieval-based methods.}
In parallel, regional geolocalization has been dominated by retrieval (either same-view or cross-view) between a query image and a geo-referenced gallery. Earlier approaches were CNN\cite{lecun1989handwritten}-based \cite{workman2015wide,liu2019lending,hu2018cvm,regmi2019cross,regmi2019bridging,toker2021coming,shi2020looking}, and the advent of Vision Transformers (ViT) \cite{dosovitskiy2020image} led to transformer-based cross-view models \cite{zhu2022transgeo,mi2024congeo,deuser2023sample4geo,yang2021cross,zhu2021vigor}. While highly precise at regional scales, scaling image-to-image retrieval to a worldwide setting is computationally prohibitive and requires unrealistically dense gallery coverage.

\begin{figure*}[t]
  \centering
  \includegraphics[width = 0.82\linewidth]{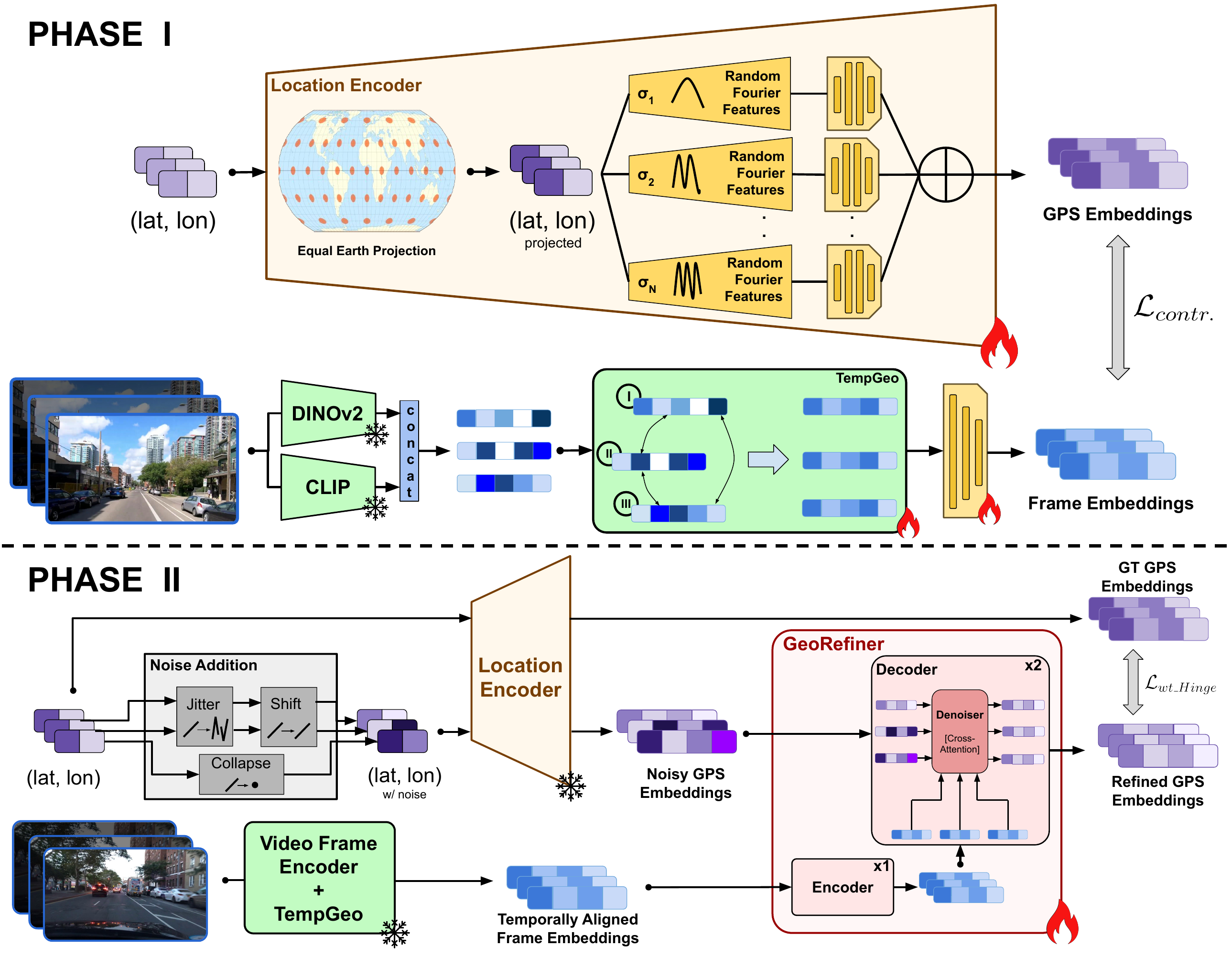}
  \caption{\textbf{Schematic Illustration of the proposed Approach.} The training of the model is divided in two phases. \textbf{(a) Phase I training.} Each frame is simultaneously passed through DINOv2 and CLIP, and concatenated to generate an embedding. TempGeo module specifically instills positional order in frames of a video and aligns them with other frames, inculcating temporal consistency within frame embeddings. The Video Frame Encoder and TempGeo module are contrastively trained with the Location Encoder. \textbf{(b) Phase II training.} The GeoRefiner module encodes the temporally aligned frame embeddings of a video. Noisy GPS embeddings are input into the decoder along with the encoded frame embeddings. GPS embeddings are cross-attended with the frame embeddings, which is trained to denoise.}
  \label{fig:arch}
  \vspace{-1em}
\end{figure*}

\subsection{Video Geolocalization}
\label{sec:related-vg}
\textbf{Early trajectory estimation.}
Vaca et al.\ \cite{vaca2012city} addressed trajectory prediction for a moving camera using Bayesian tracking, predating deep learning approaches.

\noindent \textbf{Frame-based and cross-view deep methods.}
GTFL \cite{regmi2021video} introduced a VGG \cite{simonyan2014very}-based hybrid with self-attention\cite{vaswani2017attention} to mitigate temporal inconsistency for \emph{same-view}, frame-based geolocalization. Extending to \emph{cross-view}, GAMa-Net \cite{vyas2022gama} used a hybrid backbone (ResNet \cite{he2016deep}/R3D \cite{6165309}/ViT \cite{dosovitskiy2020image}) with multi-stage hierarchical retrieval. GAReT \cite{pillai2024garet} proposed a fully transformer-based architecture with similar multi-stage retrieval and a TransRetriever module that further addresses temporal inconsistency. These methods remain confined to a few regions.

\noindent \textbf{City-level prediction and temporal aggregation.}
CityGuessr \cite{kulkarni2024cityguessr} moved toward global video geolocalization by predicting the \emph{city} for whole videos and releasing CityGuessr68k. However, it treats each video holistically rather than localizing frames, leaving temporal inconsistency unaddressed. Complementary work studies temporal aggregation to fuse framewise features into coherent video-level representations \cite{facil2019condition,garg2021seqnet,mereu2022learning}.

We adopt a \emph{Frame-to-GPS retrieval} paradigm as a feasible approach for global scale framewise video geolocalization. We aim to harness the complementary strengths of CLIP and DINOv2, and address the temporal inconsistency issue with our proposed \emph{TempGeo} and \emph{GeoRefiner} modules. 

\section{Method}
\label{sec:method}

Given a video, our goal is to predict the GPS location for each frame and thus map its trajectory. We frame this as a frame-to-GPS retrieval problem. Our approach consists of four components: Dual Frame Encoder (Section \ref{sec:dual}), TempGeo (Section \ref{sec:tempgeo}), Location Encoder (Section \ref{sec:locenc}), and GeoRefiner (Section \ref{sec:georefiner}). Combined frame embeddings from CLIP and DINOv2 encoders are temporally aligned by TempGeo. GPS coordinates are encoded via Location Encoder \cite{vivanco2024geoclip}, and contrastively trained with frame features 
GeoRefiner further refines GPS embeddings using an encoder-decoder architecture (Fig. \ref{fig:arch}).

\subsection{Dual Frame Encoder}
\label{sec:dual}

We use two complementary vision encoders to construct per-frame descriptors. CLIP provides language-aligned semantics learned from large-scale image-text pretraining \cite{radford2021learning}, which helps disambiguate landmarks, signage, and scene context. DINOv2 supplies robust self-supervised features \cite{oquab2023dinov2} that capture global appearance and are less sensitive to domain shifts. Combining them yields embeddings that are both semantically informative and visually descriptive, which benefits frame-to-GPS retrieval.

For each frame, let $\mathbf{Z}^{\text{clip}}$ and $\mathbf{Z}^{\text{dino}}$ be the token sequences produced by the two ViTs. We obtain single-vector descriptors by using the class token:
\[
\mathbf{f}_{\text{clip}}\in\mathbb{R}^{d_{\text{clip}}},\qquad
\mathbf{f}_{\text{dino}}\in\mathbb{R}^{d_{\text{dino}}}.
\]


We fuse the modalities by concatenating the two vectors:
\[
\mathbf{z}_t = \bigl[\mathbf{f}_{\text{clip}}\;\|\;\mathbf{f}_{\text{dino}}\bigr]\in\mathbb{R}^{d_{\text{clip}}+d_{\text{dino}}}
\]

The fused representation $\mathbf{z}_t$ is the frame representation consumed by TempGeo (Sec.~\ref{sec:tempgeo}). 

\subsection{TempGeo}
\label{sec:tempgeo}

Existing frame-wise video geolocalization models suffer from temporal inconsistency, as independent frame predictions can drift or include outliers that distort the trajectory. To address this, we introduce \textbf{TempGeo}, a lightweight Transformer encoder that performs learned temporal alignment through full self-attention across video frames. Unlike post-hoc smoothing or temporal pooling, TempGeo explicitly models \emph{long-range frame dependencies}, allowing uncertain or ambiguous frames to leverage contextual cues from both nearby and distant views. Given the fused, unit-normalised embeddings $\mathbf{z}_t\in\mathbb{R}^{D}$ from Sec.~\ref{sec:dual}, we add temporal positional embeddings to encode frame order,
\[
\hat{\mathbf{z}}_t=\mathbf{z}_t+\mathbf{p}_t,
\]
and process the sequence $\hat{\mathbf{Z}}=[\hat{\mathbf{z}}_1,\ldots,\hat{\mathbf{z}}_T]$ with a multi-head self-attention encoder to obtain attended embeddings $\mathbf{z}^\star_t\in\mathbb{R}^{D}$. Allowing each frame to attend to every other frame lets ambiguous views borrow context from both nearby and distant frames, and pulls isolated outliers closer to the consensus in feature space. TempGeo preserves the hidden dimensionality $D$ and uses standard pre-normalization and dropout, so it integrates cleanly with the rest of the pipeline.

TempGeo is trained jointly in Phase~I by replacing $\mathbf{z}_t$ with $\mathbf{z}^\star_t$ when computing the similarity matrix for the contrastive loss (Sec.~\ref{sec:loss}). This performs alignment \emph{before} retrieval so cross-frame context shapes the learning signal rather than relying on post-hoc smoothing. In Phase~II, TempGeo is frozen and its outputs serve as the visual input to the GeoRefiner encoder (Sec.~\ref{sec:georefiner}). This design introduces end-to-end temporal alignment within the contrastive learning framework, enabling the model to exploit non-local visual cues such as recurring landmarks or illumination changes that may span many frames, without relying on GPS or external supervision.

\subsection{Location Encoder}
\label{sec:locenc}
The Location Encoder builds upon GeoCLIP. GPS coordinates are standardized using Equal Earth Projection \cite{vsavrivc2019equal}, then represented hierarchically via Random Fourier Features (RFF) \cite{tancik2020fourier}. Each RFF is passed through separate MLPs; outputs are summed elementwise to generate final GPS embeddings, which are finetuned contrastively.

\begin{figure}
    \centering
    \includegraphics[width = 0.48\textwidth]{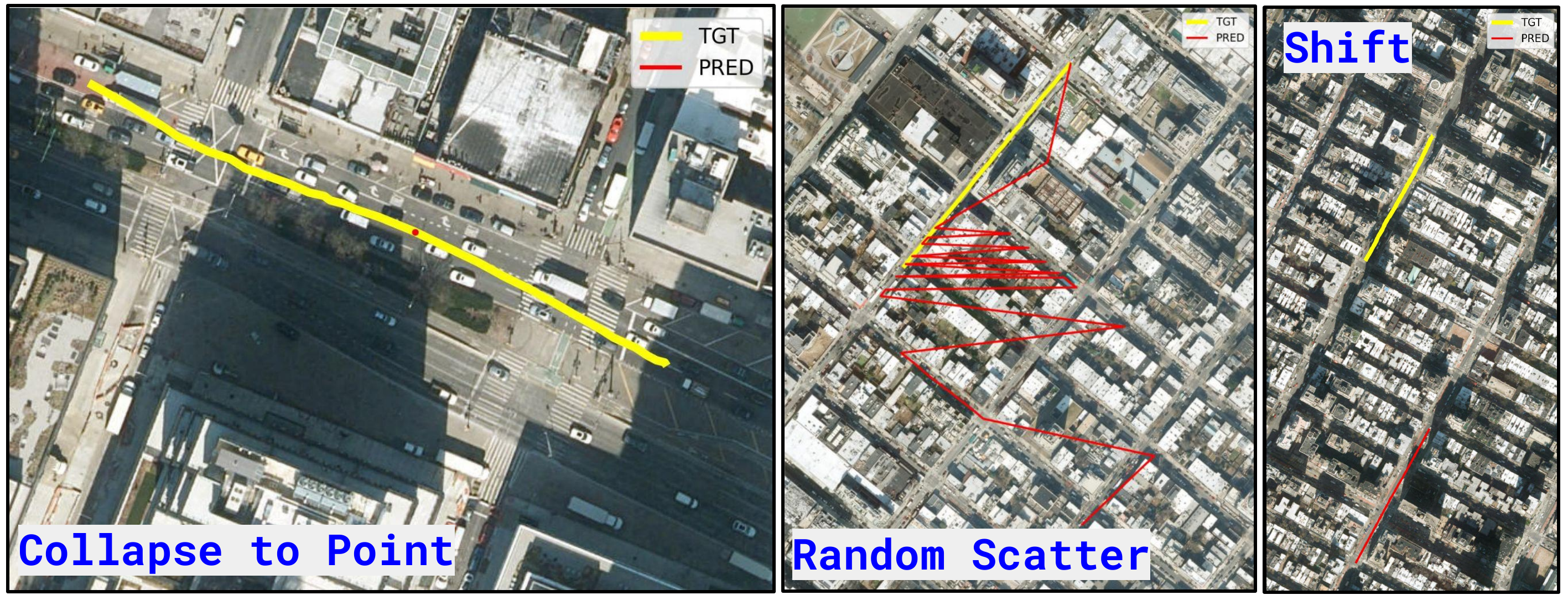}
    \vspace{-1.5em}
    \caption{\textbf{Phase I failure modes} Most commonly observed noise patterns in Phase I predictions. Phase II training refines GPS predictions to address these issues.  Yellow: GT, Red: Predicted.}
    \label{fig:mist}
    \vspace{-1.5em}
\end{figure}

\begin{figure*}[t]
  \centering
  \includegraphics[width =0.95\textwidth]{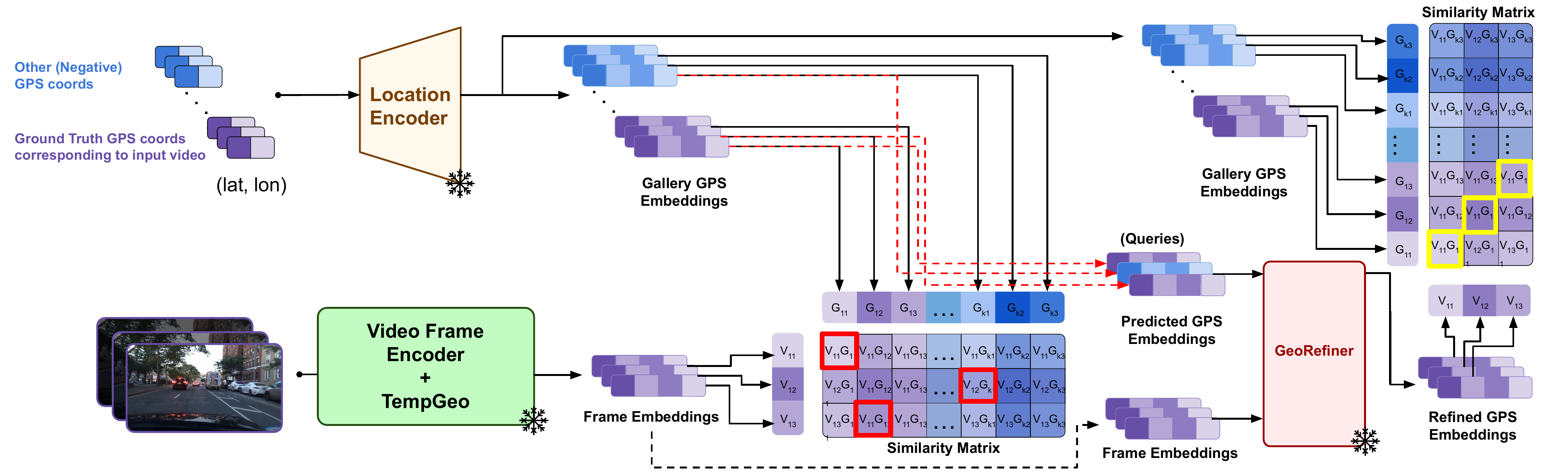}
  \vspace{-1em}
  \caption{\textbf{Inference Procedure.} Frames are passed through the dual encoders followed by TempGeo to generate frame-wise embeddings, while GPS embeddings are obtained from a gallery of coordinates via the location encoder. The frame-wise embeddings are compared  with GPS embeddings to make initial predictions. These predictions, along with corresponding frame embeddings, are refined by the GeoRefiner model, which outputs refined GPS embeddings. A second retrieval with refined embeddings produces final predictions.}

  \vspace{-1em}
 
  \label{fig:inf}
\end{figure*}

\subsection{GeoRefiner}
\label{sec:georefiner}

Even though TempGeo aligns frame features within a video, the association between visual embeddings and their corresponding GPS embeddings can remain noisy. We therefore introduce \textbf{GeoRefiner}, which refines the GPS sequence using an encoder–decoder Transformer inspired by machine translation \cite{vaswani2017attention,wang2020fairseq,lewis2019bart}. The decoder receives GPS embeddings as queries while the encoder processes the temporally aligned frame embeddings, which are then passed to the decoder as context. Cross-attention in the decoder aligns the GPS sequence to the visual tokens.

Let $\mathbf{z}_t\in\mathbb{R}^{D}$ be the fused, unit-normalised frame embedding (Sec.~\ref{sec:dual}) and $\mathbf{z}^{\star}_t$ the TempGeo output (Sec.~\ref{sec:tempgeo}). Let $\mathbf{g}_t$ denote the GPS embedding from the Location Encoder (Sec.~\ref{sec:locenc}). GeoRefiner applies cross attention over the GPS sequence in the decoder together with  the encoder output produced from $\{\mathbf{z}^{\star}_t\}$. We do not use a causal mask, allowing each GPS token to attend to the entire sequence and to all visual frames through cross-attention. The model width matches TempGeo so the module integrates cleanly with the rest of the pipeline.

To train GeoRefiner as a denoiser, we do not feed it the raw Phase~I predictions. Instead, we corrupt ground-truth GPS coordinates to simulate characteristic Phase~I failure modes observed in Fig.~\ref{fig:mist}, including sequence-wide shifts, collapses, and random jitter, and obtain their embeddings with the Location Encoder. These noisy GPS embeddings serve as decoder queries, while the corresponding $\mathbf{z}^{\star}_t$ serve as encoder inputs. The decoder outputs refined GPS embeddings $\mathbf{g}'_t$, which are aligned with the ground-truth embeddings $\mathbf{g}_t$ using the weighted Hinge loss from Sec.~\ref{sec:loss}. This training encourages the model to undo realistic noise patterns by leveraging visual context.

At inference (Fig. \ref{fig:inf}), decoder queries are formed from Phase~I GPS predictions (embedded by the Location Encoder), and the encoder receives the TempGeo features $\mathbf{z}^{\star}_t$. A single forward pass yields refined embeddings $\mathbf{g}'_t$, which are then compared against the gallery in a same-domain GPS-to-GPS retrieval to obtain the final per-frame coordinates. Using a shallow encoder–decoder keeps the added latency low while delivering trajectory improvements.

\subsection{Losses}
\label{sec:loss}
We use two losses: Phase~I employs contrastive learning, and Phase~II optimizes alignment. In Phase~I, the similarity matrix is formed by multiplying frame and GPS embeddings; the contrastive loss is the cross-entropy between this matrix and the identity. Let $V$ stack the attended frame embeddings from TempGeo, i.e., $V=\big[\mathbf{z}^\star_t\big]$ over all frames in the batch (Sec.~\ref{sec:tempgeo}), and let $G$ stack the corresponding GPS embeddings from the Location Encoder, i.e., $G=\big[\mathbf{g}_t\big]$. With these definitions:

\vspace{-1.5em}

\begin{align}
    \mathcal{L}_{contr.}(V, G) = CE(VG^T, I)
\end{align}
where $I$ is the identity. Because $\mathbf{z}_t$ are unit-normalised in Sec.~\ref{sec:dual} and TempGeo preserves dimensionality, the inner products in $VG^\top$ correspond to cosine similarity up to the GPS embedding scale.

Phase~II focuses on alignment using a weighted Hinge loss. Here, $G'$ denotes the refined GPS embeddings produced by GeoRefiner ($G'=\big[\mathbf{g'}_t\big]$), and $G$ are the ground-truth GPS embeddings from the Location Encoder. The frame-wise and video-wise components use loss matrices built from mean-squared error between similarity matrices (constructed as in Phase~I) and identity. The final objective is the weighted sum of the means of negative pairs (upper-triangular and lower-triangular) and positive pairs (diagonal elements). $\alpha$ is the weight of the negative pairs and $\beta$ of the positive pairs. We choose the weights empirically through ablations (Supplementary material Section \ref{sec:abl-alphabeta}). 

\noindent Given $G'$ and $G$, 
\begin{align}
 & \mathbf{M}_{f} = MSE(G'G^T, I); \mathbf{M}_{v} = MSE(G'_{seq}G_{seq}^T, I) \\
    \begin{split}
 &\mathcal{L}_{f} =  \alpha\left[\overline{tr_U(\mathbf{M}_{f})}+\overline{tr_L(\mathbf{M}_{f})}\right]+\beta.\overline{dia(\mathbf{M}_{f})}
    \end{split}\\
    \begin{split}
 &\mathcal{L}_{v} = \alpha\left[\overline{tr_U(\mathbf{M}_{v})}+\overline{tr_L(\mathbf{M}_{v})}\right]+\beta.\overline{dia(\mathbf{M}_{v})}
    \end{split}\\
 & \mathcal{L}_{wt\_Hinge}(G, G') =  \mathcal{L}_{f} + \mathcal{L}_{v},
\end{align}
where $G_{seq}$ and $G'_{seq}$ denote the same embeddings organized at the video level, $I$ is the identity, $\mathbf{M}_{f}$ and $\mathbf{M}_{v}$ are the frame- and video-wise loss matrices, $tr_U$ and $tr_L$ are the sets of upper- and lower-triangular elements, $dia$ the diagonal elements, and $\overline{X}$ the mean of elements in matrix $X$. This formulation ties directly to the definitions in Secs.~\ref{sec:dual}–\ref{sec:georefiner} and promotes  alignment at both frame \& video granularity.
\section{Experiments}
\label{sec:exp}
We perform experiments with multiple datasets and settings. The data we use, and the training and validation setup in regards to the same is described below. Training Details are in Supplementary material Sec. \ref{sec:td}.

\subsection{Training and Validation Data}
\label{sec:data}
We show the performance of our model and further analysis primarily on the Mapillary(MSLS)\cite{warburg2020mapillary} dataset as it is the largest image sequence dataset with frame-wise GPS annotations as well as a decent geographic coverage. As the MSLS dataset was primarily introduced for same-view frame-to-frame/sequence-to-sequence retrieval, it comprises of predivided query and database sequences. As the same-view database is not relevant for our purpose, we include all the sequences from both sets, and split it into training and validation data following the procedure described in \cite{kulkarni2024cityguessr}. We train both phases of our network on the train split, and validate on the val split with a val gallery. We show additional results on the GAMa\cite{vyas2022gama}(videos from BDD100k\cite{yu2020bdd100k}, data split from GAMa) for demonstration of the efficacy of our model on data with dense geographic coverage. We also show our performance on an alternate task of city prediction on CityGuessr68k and compare with the current state-of-the-art as well as foundation models.

\subsection{Uniform Grid Gallery}
\label{sec:gallery}
Framewise retrieval requires a reference gallery, which in our case comprises of GPS coordinates. To make sure that the retrieval is fair, the model should not have access to any information pertaining to the validation set. Thus we construct a uniform grid gallery using the training data, such that model has to perform a completely blind retrieval. The gallery construction procedure for all datasets is described in detail in the Supplementary material Sec. \ref{sec:refgall}.

\subsection{Evaluation Metrics}
\label{sec:metric}
We show our results on a variety of metrics. As standard practice with worldwide geolocalization models, we show accuracy results on certain distance thresholds. As we observe that our model performs very well($\sim98\%$) on and beyond city-level(25km), we do not show performance for over 25km. Instead for additional granularity, we add finer thresholds of 500m(Sub-street level) and 5km(Locality level) to the 2 standard ones of 1km and 25km resulting in a total of 4 thresholds. We also measure the model's precision in terms of median distance error.

As frame-wise performance does not accurately measure the quality of sequences generated from the frame-wise predictions, we also show our performance on a diverse set of video metrics. First, we show results on accuracy with distance thresholds for videos (computation procedure described in detail in Supplementary material Sec. \ref{sec:vidacc}).

Then, for analyzing the quality of trajectories, we show performance on additional metrics of Discrete Frechet Distance(DFD)\cite{eiter1994computing} and Mean Range Difference(MRD)\cite{gaharwar2023xi}, adapted for 2D sequences. We adapt MRD for 2D by computing MRD separately for latitude/longitude sequences (More details in Supplementary material Sec. \ref{sec:dfd} and \ref{sec:mrd}).
\section{Results and Discussion}
\label{sec:res}
We present the results of the experiments described in Section~\ref{sec:exp}.
We first describe the baselines and state-of-the-art methods used for
comparison, show qualitative results, then report performance on MSLS, GAMa, and CityGuessr68k, followed by an analysis of the accuracy--throughput tradeoff and the effect of gallery resolution.

\begin{table*}[t]
\caption{Comparison of our model with baseline methods on Mapillary (MSLS) dataset}
  \vspace{-1em}
  \setlength{\tabcolsep}{1.5pt}
    \renewcommand{\arraystretch}{0.9}
\resizebox{\linewidth}{!}{
\begin{tabular}{l|ccccc|ccccccc}
\hline \hline
\multicolumn{1}{c|}{\multirow{4}{*}{\textbf{Model}}} & \multicolumn{5}{c|}{\textbf{Frame-wise Metrics}} & \multicolumn{7}{c}{\textbf{Video Metrics}}                                                                                                                                                                                                                                                                      \\ \cline{2-13} 
\multicolumn{1}{c|}{} & \multicolumn{4}{c|}{\textbf{Distance ($a_r$ {[}\%{]} @ km)} $\uparrow$} & \multirow{3}{*}{\textbf{\begin{tabular}[c]{@{}c@{}}Median\\ Distance\\ Error(km) $\downarrow$\end{tabular}}} & \multicolumn{4}{c|}{\textbf{Distance ($a_r$ {[}\%{]} @ km)} $\uparrow$} & \multicolumn{1}{c|}{\multirow{3}{*}{\textbf{\begin{tabular}[c]{@{}c@{}}Median\\ Distance\\ Error(km) $\downarrow$\end{tabular}}}} & \multicolumn{1}{c|}{\multirow{3}{*}{\textbf{DFD} $\downarrow$}} & \multirow{3}{*}{\textbf{MRD} $\downarrow$} \\
\multicolumn{1}{c|}{} & \textbf{Sub-street} & \textbf{Street} & \textbf{Locality} & \multicolumn{1}{c|}{\textbf{City}} & & \textbf{Sub-street} & \textbf{Street} & \textbf{Locality} & \multicolumn{1}{c|}{\textbf{City}} & \multicolumn{1}{c|}{} & \multicolumn{1}{c|}{} &                               \\
\multicolumn{1}{c|}{} & \textbf{0.5 km (500 m)} & \textbf{1 km} & \textbf{5 km} & \multicolumn{1}{c|}{\textbf{25 km}} & & \textbf{0.5 km (500 m)} & \textbf{1 km} & \textbf{5 km} & \multicolumn{1}{c|}{\textbf{25 km}} & \multicolumn{1}{c|}{} & \multicolumn{1}{c|}{} &                               \\ \hline
GeoCLIP\cite{vivanco2024geoclip}-ZeroShot & 0.9 & 2.7 & 22.9 & \multicolumn{1}{c|}{84.6} & 11.54 & 1.3 & 3.8 & 27.6 & \multicolumn{1}{c|}{89.9} & \multicolumn{1}{c|}{9.85} & \multicolumn{1}{c|}{24.94} & 2.83                          \\ \hline
PlaNet\cite{weyand2016planet} & 4.4 & 10.5 & 38.4 & \multicolumn{1}{c|}{81.0} & 7.38 & 3.9 & 8.6 & 35.6 & \multicolumn{1}{c|}{78.9} & \multicolumn{1}{c|}{7.66} & \multicolumn{1}{c|}{67.42} & 9.86                          \\
ISNs\cite{muller2018geolocation} & 4.9 & 11.2 & 38.6 & \multicolumn{1}{c|}{84.3} & 7.22 & 4.3 & 10.1 & 36.0 & \multicolumn{1}{c|}{80.4} & \multicolumn{1}{c|}{7.51} & \multicolumn{1}{c|}{62.55} & 8.95                          \\
GeoDecoder\cite{clark2023we} & 6.3 & 14.7 & 50.6 & \multicolumn{1}{c|}{88.5} & 4.9 & 5.4 & 13.5 & 46.6 & \multicolumn{1}{c|}{88.1} & \multicolumn{1}{c|}{5.58} & \multicolumn{1}{c|}{54.16} & 5.07                          \\ \hline
CLIP\cite{radford2021learning} (Classification) & 6.1 & 14.3 & 49.7 & \multicolumn{1}{c|}{92.7} & 5.05 & 6.1 & 13.6 & 50.3 & \multicolumn{1}{c|}{95.4} & \multicolumn{1}{c|}{4.97} & \multicolumn{1}{c|}{25.64} & 1.96                          \\
DINOv2\cite{oquab2023dinov2} (Classification) & 7.4 & 18.1 & 58.2 & \multicolumn{1}{c|}{96.8} & 3.86 & 7.9 & 18.4 & 60.8 & \multicolumn{1}{c|}{96.2} & \multicolumn{1}{c|}{3.62} & \multicolumn{1}{c|}{4.28} & 1.60                           \\ \hline
GeoCLIP\cite{vivanco2024geoclip}-FineTuned & 8.3 & 22.5 & 63.0 & \multicolumn{1}{c|}{93.9} & 2.97 & 6.7 & 18.6 & 58.2 & \multicolumn{1}{c|}{94.1} & \multicolumn{1}{c|}{3.69} & \multicolumn{1}{c|}{22.52} & 2.82                          \\ \hline
\textbf{VidTAG (Ours)} & \textbf{21.5} & \textbf{41.0} & \textbf{76.7} & \multicolumn{1}{c|}{\textbf{97.9}} & \textbf{1.35} & \textbf{22.2} & \textbf{39.8} & \textbf{74.3} & \multicolumn{1}{c|}{\textbf{97.5}} & \multicolumn{1}{c|}{\textbf{1.49}} & \multicolumn{1}{c|}{\textbf{3.87}} & \textbf{1.07}                 \\ \hline \hline
\end{tabular}
}
\vspace{-0.5em}
\label{tab:msls}
\end{table*}

\begin{table*}[t]
\caption{Comparison of our model with baseline methods on GAMa split of the BDD100k dataset}
  \vspace{-1em}
  \setlength{\tabcolsep}{1.5pt}
    \renewcommand{\arraystretch}{0.95}
\resizebox{\linewidth}{!}{
\begin{tabular}{l|ccccc|ccccccc}
\hline\hline
\multicolumn{1}{c|}{\multirow{4}{*}{\textbf{Model}}} & \multicolumn{5}{c|}{\textbf{Frame-wise Metrics}} & \multicolumn{7}{c}{\textbf{Video Metrics}}                                                                                                                                                                                                                                                                      \\ \cline{2-13} 
\multicolumn{1}{c|}{} & \multicolumn{4}{c|}{\textbf{Distance ($a_r$ {[}\%{]} @ km)} $\uparrow$} & \multirow{3}{*}{\textbf{\begin{tabular}[c]{@{}c@{}}Median\\ Distance\\ Error(km) $\downarrow$\end{tabular}}} & \multicolumn{4}{c|}{\textbf{Distance ($a_r$ {[}\%{]} @ km) $\uparrow$}} & \multicolumn{1}{c|}{\multirow{3}{*}{\textbf{\begin{tabular}[c]{@{}c@{}}Median\\ Distance\\ Error(km) $\downarrow$\end{tabular}}}} & \multicolumn{1}{c|}{\multirow{3}{*}{\textbf{DFD} $\downarrow$}} & \multirow{3}{*}{\textbf{MRD} $\downarrow$} \\
\multicolumn{1}{c|}{} & \textbf{Sub-street} & \textbf{Street} & \textbf{Locality} & \multicolumn{1}{c|}{\textbf{City}} & & \textbf{Sub-street} & \textbf{Street} & \textbf{Locality} & \multicolumn{1}{c|}{\textbf{City}} & \multicolumn{1}{c|}{} & \multicolumn{1}{c|}{} &                               \\
\multicolumn{1}{c|}{} & \textbf{0.5 km} & \textbf{1 km} & \textbf{5 km} & \multicolumn{1}{c|}{\textbf{25 km}} & & \textbf{0.5 km} & \textbf{1 km} & \textbf{5 km} & \multicolumn{1}{c|}{\textbf{25 km}} & \multicolumn{1}{c|}{} & \multicolumn{1}{c|}{} &                               \\ \hline
GeoCLIP\cite{vivanco2024geoclip}-ZeroShot & 4.1 & 10.1 & 33.5 & \multicolumn{1}{c|}{74.9} & 11.15 & 4.0 & 9.6 & 32.2 & \multicolumn{1}{c|}{75.3} & \multicolumn{1}{c|}{11.08} & \multicolumn{1}{c|}{14.29} & 1.51            \\ \hline
PlaNet\cite{weyand2016planet} & 7.7 & 15.4 & 44.1 & \multicolumn{1}{c|}{84.5} & 6.26 & 6.3 & 14.3 & 43.5 & \multicolumn{1}{c|}{86.4} & \multicolumn{1}{c|}{6.21} & \multicolumn{1}{c|}{43.19} & 6.76                          \\
ISNs\cite{muller2018geolocation} & 8.2 & 16.3 & 44.8 & \multicolumn{1}{c|}{86.1} & 6.11 & 7.0 & 14.9 & 45.1 & \multicolumn{1}{c|}{87.2} & \multicolumn{1}{c|}{6.06} & \multicolumn{1}{c|}{40.07} & 6.14                          \\
GeoDecoder\cite{clark2023we} & 12.9 & 23.2 & 52.8 & \multicolumn{1}{c|}{88.9} & 4.44 & 10.7 & 20.4 & 52.8 & \multicolumn{1}{c|}{87.7} & \multicolumn{1}{c|}{4.50} & \multicolumn{1}{c|}{28.46} & 2.92                          \\ \hline
CLIP\cite{radford2021learning} (Classification) & 10.9 & 20.3 & 49.9 & \multicolumn{1}{c|}{90.2} & 5.03 & 9.4 & 18.4 & 50.6 & \multicolumn{1}{c|}{90.3} & \multicolumn{1}{c|}{4.90} & \multicolumn{1}{c|}{17.34} & 1.26                          \\
DINOv2\cite{oquab2023dinov2} (Classification) & 13.2 & 23.2 & 50.4 & \multicolumn{1}{c|}{89.7} & 4.92 & 11.5 & 21.0 & 52.5 & \multicolumn{1}{c|}{90.3} & \multicolumn{1}{c|}{4.50} & \multicolumn{1}{c|}{16.90} & 1.85                          \\ \hline
GeoCLIP\cite{vivanco2024geoclip}-FineTuned & 16.3 & 28.3 & 57.8 & \multicolumn{1}{c|}{89.1} & 3.28 & 15.9 & 28.7 & 61.2 & \multicolumn{1}{c|}{90.9} & \multicolumn{1}{c|}{2.87} & \multicolumn{1}{c|}{6.50} & 0.50                          \\ \hline
\textbf{VidTAG (Ours)} & \textbf{35.4} & \textbf{53.1} & \textbf{77.8} & \multicolumn{1}{c|}{\textbf{94.4}} & \textbf{0.88} & \textbf{35.8} & \textbf{53.6} & \textbf{78.4} & \multicolumn{1}{c|}{\textbf{94.6}} & \multicolumn{1}{c|}{\textbf{0.86}} & \multicolumn{1}{c|}{\textbf{0.39}} & \textbf{0.17}                 \\ \hline\hline
\end{tabular}
}
\vspace{-1em}
\label{tab:GAMa}
\end{table*}

\subsection{Baselines and SoTA Methods}
\label{sec:basesota}
As frame-to-GPS video geolocalization has not been studied before, we compare
to state-of-the-art image geolocalization models and additional baselines. Our main baseline is GeoCLIP\cite{vivanco2024geoclip}, the image equivalent of our approach and a state-of-the-art method for worldwide image geolocalization. We report both zero-shot and fine-tuned variants for a fair comparison. We also evaluate classification-based models PlaNet~\cite{weyand2016planet}, ISNs~\cite{muller2018geolocation}, and GeoDecoder~\cite{clark2023we}, which is a strong SoTA geolocation model. All models are trained on the same training data, and perform classification over S2 cells generated with the S2 geometry library\footnote{\url{code.google.com/archive/p/s2-geometry-library}}. The training and validation procedures for these baselines are described in Supplementary Sec.~\ref{sec:sota}. In addition, we fine-tune classification models using CLIP and DINOv2 backbones under similar settings; these are used for comparison on MSLS and GAMa. For CityGuessr68k, we compare to the models evaluated in the original paper, as well as CLIP-based geo-foundation models
\cite{astruc2024openstreetview,haas2023learning,vivanco2024geoclip} and
multi-modal large language model (MLLM) approaches
\cite{li2024georeasoner,bai2025qwen2}.

\begin{figure}[t]
  \centering
  \includegraphics[width = 0.4\textwidth]{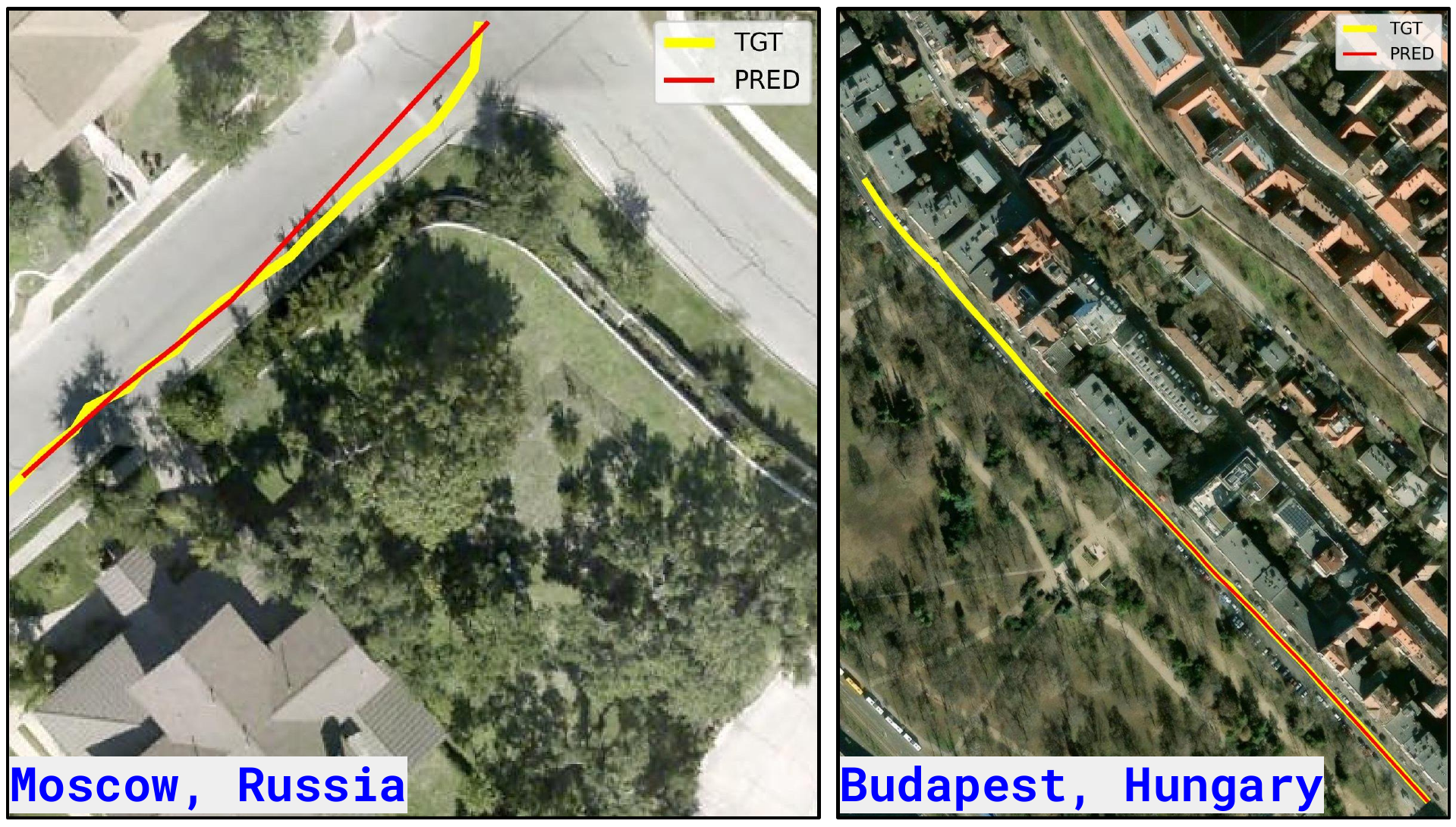}
  \vspace{-0.9em}
  \caption{\textbf{Qualitative Results.} Mapped near-perfect geolocalized trajectories predicted by VidTAG. Yellow: GT, Red: Predicted.}
  \vspace{-1em}
  \label{fig:qual}
\end{figure}

\subsection{Qualitative Results}
\label{sec:qual}
The primary objective of video geolocalization is mapping the trajectory of a given video. Thus we show qualitative performance of our model by plotting a trajectories.  Fig. \ref{fig:qual} shows two trajectories traced by predictions of our model in different countries from MSLS. The yellow curves represent the ground truth trajectories constructed from the given GPS annotations, while red curves depict the trajectories formed by the GPS coordinates predicted by the model. Additional Qualitative Analysis (using videos from GAMa) is provided in Supplementary material Sec. \ref{sec:qualgama}.

\subsection{Quantitative Results}
\label{sec:quant}
\noindent\textbf{MSLS:}
Table~\ref{tab:msls} shows results on the MSLS dataset. Our model outperforms
the best baseline by a clear margin across all frame-wise and video metrics.
In particular, VidTAG improves by roughly 20\% over fine-tuned GeoCLIP at
the 1\,km threshold. DINOv2 (Classification) and GeoCLIP-FineTuned perform
well at coarser thresholds (5\,km, 25\,km), but struggle in the finer regime
(0.5\,km, 1\,km) where our model excels. Trajectory quality is also
substantially better, as indicated by lower DFD and MRD, highlighting the
effectiveness of TempGeo and GeoRefiner.

\begin{table}[t]
\caption{City-level prediction results on CityGuessr68k}
  \vspace{-1em}
    \setlength{\tabcolsep}{1.5pt}
    \renewcommand{\arraystretch}{0.6}
\resizebox{0.46\textwidth}{!}{
\begin{tabular}{lcccc}
\toprule
\multicolumn{1}{c}{\textbf{Model}} & \multicolumn{1}{c}{\textbf{City $\uparrow$}} & \multicolumn{1}{c}{\textbf{State $\uparrow$}} & \multicolumn{1}{c}{\textbf{Country $\uparrow$}} & \multicolumn{1}{c}{\textbf{Continent $\uparrow$}} \\ \midrule
PlaNet\cite{weyand2016planet} & 55.8 & 56.3 & 60.8 & 74.1                                   \\
ISNs\cite{muller2018geolocation} & 59.5 & 59.9 & 64.1 & 75.9                                   \\
GeoDecoder\cite{clark2023we} & 64.2 & 64.5 & 69.5 & 79.9                                   \\ \midrule
GeoReasoner\cite{li2024georeasoner} & 38.5 & 42.8 & 64.4 & 81.9                                    \\
Qwen2.5-VL\cite{bai2025qwen2} & 55.1 & 59.9 & 78.2 &  91.3                                   \\ \midrule
OSV\cite{astruc2024openstreetview} & 23.1 & 33.5 & 55.6 &  78.1                                   \\
StreetCLIP\cite{haas2023learning} & 55.6 & 56.8 & 69.7 &  91.2                                   \\
GeoCLIP\cite{vivanco2024geoclip} & 57.8 & 60.5 & 75.9 & 90.8                   \\  \midrule
Timesformer\cite{bertasius2021space} & 60.9 & 61.4 & 66.1 & 78.4                                   \\
VideoMAE\cite{tong2022videomae} & 64.5 & 64.5 & 65.9 & 74.4                                   \\
CityGuessr\cite{kulkarni2024cityguessr} & 69.6 & 70.2 & 74.8 & 83.8                                   \\ 
\midrule
\textbf{VidTAG(Ours)} & \multicolumn{1}{c}{\textbf{94.9}} & \multicolumn{1}{c}{\textbf{95.5}} & \multicolumn{1}{c}{\textbf{96.8}} & \multicolumn{1}{c}{\textbf{98.5}}                   \\ \bottomrule
\end{tabular}
}
\vspace{-1.7em}
\label{tab:cg}
\end{table}

\noindent\textbf{GAMa:}
We next evaluate on the framewise annotated GAMa~\cite{vyas2022gama} dataset
derived from BDD100k~\cite{yu2020bdd100k}. As shown in Table~\ref{tab:GAMa},
VidTAG again shows a substantial improvement over fine-tuned GeoCLIP, with
an increase of almost 25\% at the 1\,km threshold. The very low DFD and MRD
scores demonstrate that our model produces high-quality trajectories in
fine-grained, densely sampled scenarios.

\noindent\textbf{CityGuessr68k:}
CityGuessr68k~\cite{kulkarni2024cityguessr} is a global video dataset covering
166 cities with hierarchical geographical labels. We adapt it to a
city-level GPS retrieval task, where the gallery contains the GPS coordinates
of the city centres. We compare VidTAG to image classification, MLLM,
CLIP-based, and video classification baselines. As shown in
Table~\ref{tab:cg}, our model outperforms all previous state-of-the-art
approaches by more than 25\% at the city level, reaching $\sim$95\% accuracy.
It also achieves the best performance at state, country, and continent levels,
indicating strong generalization to global-scale datasets. A more detailed
analysis of hard cases is provided in Supplementary Sec.~\ref{sec:city}.

\begin{table*}[t]
\caption{Component-wise Ablation Study}
  \vspace{-1em}
  \setlength{\tabcolsep}{1.5pt}
    \renewcommand{\arraystretch}{0.95}
\resizebox{\linewidth}{!}{
\begin{tabular}{cccc|ccc|ccccc}
\hline \hline
\multicolumn{4}{c|}{\multirow{2}{*}{\textbf{Ablation Component}}}                                                & \multicolumn{3}{c|}{\textbf{Frame-wise Metrics}}                                                                                                              & \multicolumn{5}{c}{\textbf{Video Metrics}}                                                                                                                                                                                                                              \\ \cline{5-12} 
\multicolumn{4}{c|}{}                                                                                            & \multicolumn{2}{c|}{\textbf{Distance ($a_r$ {[}\%{]} @ km)} $\uparrow$} & \multirow{3}{*}{\textbf{\begin{tabular}[c]{@{}c@{}}Median\\ Distance\\ Error(km) $\downarrow$\end{tabular}}} & \multicolumn{2}{c|}{\textbf{Distance ($a_r$ {[}\%{]} @ km)} $\uparrow$} & \multicolumn{1}{c|}{\multirow{3}{*}{\textbf{\begin{tabular}[c]{@{}c@{}}Median\\ Distance\\ Error(km) $\downarrow$\end{tabular}}}} & \multicolumn{1}{c|}{\multirow{3}{*}{\textbf{DFD} $\downarrow$}} & \multirow{3}{*}{\textbf{MRD} $\downarrow$} \\ \cline{1-4}
\multicolumn{2}{c}{\textbf{Backbone}} & \multirow{2}{*}{\textbf{TempGeo}} & \multirow{2}{*}{\textbf{GeoRefiner}} & \textbf{Street}   & \multicolumn{1}{c|}{\textbf{Locality}}  &                                                                                                 & \textbf{Street}   & \multicolumn{1}{c|}{\textbf{Locality}}  & \multicolumn{1}{c|}{}                                                                                                & \multicolumn{1}{c|}{}                              &                               \\
\textbf{CLIP}    & \textbf{DINOv2}    &                                   &                                      & \textbf{1 km}     & \multicolumn{1}{c|}{\textbf{5 km}}      &                                                                                                 & \textbf{1 km}     & \multicolumn{1}{c|}{\textbf{5 km}}      & \multicolumn{1}{c|}{}                                                                                                & \multicolumn{1}{c|}{}                              &                               \\ \hline
            \Yes     &     \No               &        \No                           &        \No                              & 22.5              & \multicolumn{1}{c|}{63.0}               & 2.97                                                                                            & 18.6              & \multicolumn{1}{c|}{58.2}               & \multicolumn{1}{c|}{3.69}                                                                                            & \multicolumn{1}{c|}{22.52}                         & 2.82                          \\
            \No     &          \Yes          &           \No                        &         \No                             &     26.4              & \multicolumn{1}{c|}{70.1}                   &     2.13                                                                                            &       24.7            & \multicolumn{1}{c|}{69.4}                   & \multicolumn{1}{c|}{2.32}                                                                                                & \multicolumn{1}{c|}{9.15}                              &           1.03                    \\
            \No     &       \Yes             &            \Yes                       &       \No                               &        30.2           & \multicolumn{1}{c|}{72.5}                   &    1.91                                                                                             &        30.5           & \multicolumn{1}{c|}{72.3}                   & \multicolumn{1}{c|}{1.99}                                                                                                & \multicolumn{1}{c|}{\textbf{3.01}}                              &         \textbf{0.41}                      \\
            \Yes     &           \Yes         &        \Yes                           &       \No                               & 40.1              & \multicolumn{1}{c|}{75.4}               & 1.38                                                                                            & 38.7              & \multicolumn{1}{c|}{73.5}               & \multicolumn{1}{c|}{1.52}                                                                                            & \multicolumn{1}{c|}{7.63}                          & \multicolumn{1}{c}{1.57}      \\ \hline
            \Yes     &      \Yes              &          \Yes                         &        \Yes                              & \textbf{41.0}              & \multicolumn{1}{c|}{\textbf{76.7}}               & \textbf{1.35}                                                                                            & \textbf{39.8}              & \multicolumn{1}{c|}{\textbf{74.3}}               & \multicolumn{1}{c|}{\textbf{1.49}}                                                                                            & \multicolumn{1}{c|}{3.87}                          & 1.07                          \\ \hline \hline
\end{tabular}
}
\vspace{-1.5em}
\label{tab:abl}
\end{table*}

\begin{figure}[t]
  \centering
  \includegraphics[width = 0.42\textwidth]{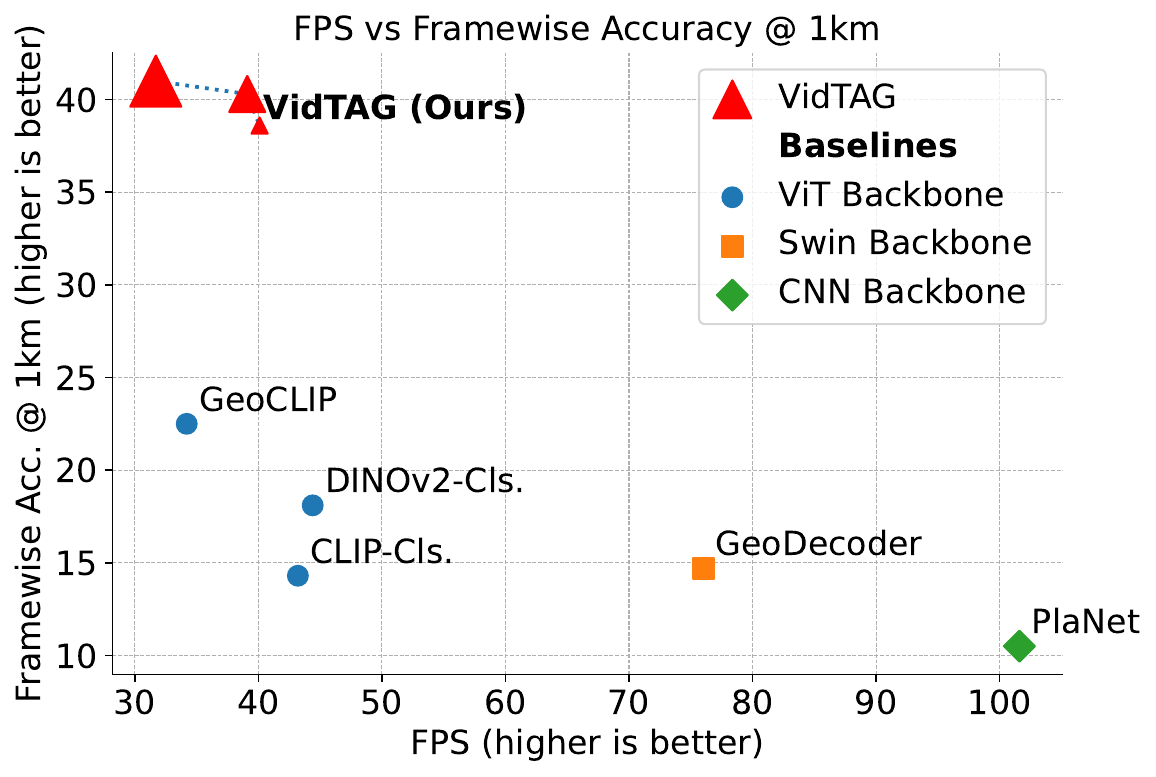}
  \vspace{-1em}
  \caption{\textbf{Performance vs Throughput Tradeoff.} Our model is significantly better with only a minimal throughput decrease.}
  \label{fig:latency}
\vspace{-2em}
\end{figure}

\subsection{Accuracy vs Throughput Tradeoff Analysis}
\label{sec:latency}
Traditional models exhibit a clear tradeoff between geolocalization accuracy
and inference throughput: higher performance usually comes with larger model
sizes and lower FPS. Figure~\ref{fig:latency} summarizes this behaviour.
CNN\cite{lecun1989handwritten}-based PlaNet\cite{weyand2016planet} achieves the fastest inference at
around 100~FPS. Swin-B\cite{liu2021swin}-based GeoDecoder\cite{clark2023we}
improves geolocalization accuracy but reduces throughput by about 24~FPS.
DINOv2\cite{oquab2023dinov2} and CLIP\cite{radford2021learning}
classification baselines are slower still, with only modest performance gains.
Retrieval-based GeoCLIP\cite{vivanco2024geoclip} further reduces FPS due to
the additional retrieval computation. Overall, these models follow a roughly
linear performance--throughput tradeoff.

In contrast, our model achieves significantly better accuracy than GeoCLIP
with only a minimal decrease in throughput, demonstrating that the proposed
modules are lightweight yet effective. Figure~\ref{fig:latency} also shows our
model evaluated with three different gallery resolutions, which are discussed
in Sec.~\ref{sec:resol}. A detailed computational cost analysis is provided in
Supplementary Sec.~\ref{sec:compute}.

\subsection{Effect of Gallery Resolution}
\label{sec:resol}
The resolution of the grid gallery has a sizable impact on both performance
and computational cost. Finer galleries provide more precise candidate
locations, but their size scales quadratically with resolution, increasing
compute and memory requirements. Table~\ref{tab:resol} compares our model on
Mapillary (MSLS)\cite{warburg2020mapillary} using galleries with resolutions
of 1\,km, 0.5\,km, and 0.1\,km. For our main evaluations, we use a resolution of 0.1\,km on MSLS, which is finer than the lowest distance threshold (0.5\,km) while keeping the gallery size manageable. Finer resolutions would lead to a massive gallery with diminishing returns. For GAMa, we choose a 0.5\,km resolution because of availability of denser data.

 We include further discussion in the Supplementary material. Sec. \ref{sec:unified} demonstrates the generalizability of our proposed techniques, with a unified model which significantly outperforms state-of-the-art methods and baselines including finetuned GeoCLIP\cite{vivanco2024geoclip} on MSLS and GAMa, and CityGuessr\cite{kulkarni2024cityguessr} on CityGuessr68k. Sec. \ref{sec:vidlen} shows our model's robustness towards video length variation.


\begin{table}[t]
\caption{Impact of Gallery Resolution}
  \vspace{-1em}
  \setlength{\tabcolsep}{1.5pt}
    \renewcommand{\arraystretch}{0.95}
\resizebox{0.48\textwidth}{!}{
\begin{tabular}{c|ccc|c|c|c}
\hline
\multirow{2}{*}{\textbf{Resolution}} & \multicolumn{3}{c|}{\textbf{Distance (a\_r {[}\%{]} @ km) $\uparrow$}} & \multirow{3}{*}{\begin{tabular}[c]{@{}c@{}}\textbf{Median}\\ \textbf{Distance}\\ \textbf{Error(km)} $\downarrow$\end{tabular}} & \multirow{3}{*}{\textbf{DFD} $\downarrow$} & \multirow{3}{*}{\textbf{MRD} $\downarrow$} \\
                                  & \textbf{Sub-street}           & \textbf{Street}      & \textbf{Locality}      &                                                                                        &                      &                      \\
                                 \textbf{(km)} & \textbf{0.5 km (500 m)}       & \textbf{1 km}        & \textbf{5 km}          &                                                                                        &                      &                      \\ \hline
1                                & 16.7                 & 38.6        & 75.1          & 1.43                                                                                   & 4.21                 & 1.18                 \\
0.5                              & 20.1                 & 40.3        & 75.5          & 1.35                                                                                   & 3.98                 & 1.11                 \\
0.1                              & 21.5                 & 41.0        & 76.7          & 1.35                                                                                   & 3.87                 & 1.07                 \\ \hline
\end{tabular}
}
\vspace{-1em}
\label{tab:resol}
\end{table}

\vspace{-0.5em}
\section{Ablation Studies}
\label{sec:abl}

\subsection{Component-wise Analysis}
\label{sec:component}
As outlined in Section \ref{sec:method}, our model comprises several components whose effects are quantified in Table \ref{tab:abl}. Beginning with a Finetuned GeoCLIP baseline that yields moderate results, we observe consistent gains as additional components are introduced. The full configuration combining CLIP and DINOv2 backbones with TempGeo and GeoRefiner delivers the strongest overall performance. 

The trajectory metrics exhibit a clear progression: replacing CLIP with DINOv2 reduces DFD and MRD, and adding TempGeo further decreases both. Reintroducing CLIP then provides a 10\% improvement in 1km accuracy through complementarity with DINOv2, albeit at the cost of some smoothness, as reflected by increases in DFD and MRD. Incorporating GeoRefiner restores smoothness, bringing these values back down. Taken together, the results indicate that TempGeo and GeoRefiner successfully improve temporal consistency, their primary objective, while the joint use of CLIP and DINOv2 drives accuracy.

\subsection{Importance of Choice in Dual Encoders}
As discussed in Sec. \ref{sec:dual}, our model employs a dual frame encoder approach, which entails combining image features from both CLIP\cite{radford2019language} and DINOv2\cite{oquab2023dinov2} to obtain the final frame feature embedding. The choice of the two encoders is not arbitrary, as both encoders complement each other and provide utility to the model, which the other one cannot. CLIP provides the semantic understanding which DINOv2 lacks, and DINOv2 provides a strong general representation that CLIP cannot provide. Thus, they overcome each others' shortcomings. We test this hypothesis by training a variation of our model in which we replace the DINOv2 encoder with a SigLIP\cite{zhai2023sigmoid} encoder, which is essentially an upgraded CLIP. Table \ref{tab:dual} shows that despite similar feature size, the CLIP-DINOv2 combination is superior in terms of performance, which empirically verifies our hypothesis.

Additional ablations examining the choice of $\alpha$ and $\beta$ hyperparameters, number of MLP layers in image encoder, $\sigma$ values for RFF in location encoder and training sequence length appear in Supplementary Sec. \ref{sec:add-abl}.

\begin{table}[t]
\caption{Effectiveness of our dual encoder}
  \vspace{-1em}
\setlength{\tabcolsep}{1.5pt}
    \renewcommand{\arraystretch}{0.95}
\resizebox{0.47\textwidth}{!}{
\begin{tabular}{c|cccc|c}
\hline
\multirow{3}{*}{\textbf{Encoders}}                             & \multicolumn{4}{c|}{\textbf{Distance ($a_r$ {[}\%{]} @ km) $\uparrow$}} & \multirow{3}{*}{\begin{tabular}[c]{@{}c@{}}\textbf{Median}\\ \textbf{Distance}\\ \textbf{Error(km)} $\downarrow$\end{tabular}} \\ 
                                                   & \textbf{Sub-street}       & \textbf{Street}         & \textbf{Locality}   & \textbf{City}    &                                                                                        \\
                                                   & \textbf{0.5 km (500 m)}   & \textbf{1 km}           & \textbf{5 km}     & \textbf{25 km}      &                                                                                        \\ \hline
\multicolumn{1}{l|}{CLIP + SigLIP}    & 11.4             & 27.8           & 71.0    & 97.1       & 2.15                                                                                   \\
\multicolumn{1}{l|}{\textbf{CLIP + DINOv2}} & \textbf{22.9}    & \textbf{40.1}  & \textbf{75.4}  & \textbf{97.4} & \textbf{1.38}                                                                          \\ \hline
\end{tabular}
}
\vspace{-1em}
\label{tab:dual}
\end{table}
\section{Conclusion}
\label{sec:conc}

We propose a framework for global video geolocalization, introducing a frame-to-GPS retrieval approach that achieves precise and scalable localization with temporally consistent trajectories. We leverage the complementary strengths of DINOv2 and CLIP for better feature representation. Our TempGeo module ensures temporal alignment, and the GeoRefiner module refines noisy GPS predictions for improved accuracy. Experiments on datasets such as MSLS, GAMa, and CityGuessr68k demonstrate state-of-the-art performance in both fine-grained localization and trajectory consistency, significantly outperforming prior works. 
\section*{Acknowledgments}

This work was supported in parts by the US Army contract W911NF-2120192 and National Geospatial-Intelligence Agency(NGA) Award \#HM0476-20-1-0001. We would like to extend our gratitude to all the reviewers for their valuable suggestions. We would also like to thank Ishan Rajendrakumar Dave, Sirnam Swetha, David G. Shatwell and Jeffrey A. Chan-Santiago for their contributions and insightful discusssions.
{
    \small
    \bibliographystyle{ieeenat_fullname}
    \bibliography{main}

@String(ECCV= {Eur. Conf. Comput. Vis.})

@String(ICIP = {IEEE Int. Conf. Image Process.})

@String(ECCV  = {ECCV})

@String(ICIP  = {ICIP})

@article{facil2019condition,
  title={Condition-invariant multi-view place recognition},
  author={Facil, J.M. and Olid, D. and Montesano, L. and Civera, J.},
  journal={arXiv preprint arXiv:1902.09516},
  year={2019}
}

@article{garg2021seqnet,
  title={SeqNet: Learning Descriptors for Sequence-Based Hierarchical Place Recognition},
  author={Garg, S. and Milford, M.},
  journal={IEEE Robotics and Automation Letters},
  volume={6},
  number={3},
  pages={4305--4312},
  year={2021},
  publisher={IEEE}
}

@inproceedings{mereu2022learning,
  title={Learning Sequential Descriptors for Sequence-Based Visual Place Recognition},
  author={Mereu, R. and others},
  booktitle={Proceedings of the IEEE International Conference on Robotics and Automation (ICRA)},
  year={2022}
}

@inproceedings{he2016deep,
  title={Deep residual learning for image recognition},
  author={He, Kaiming and Zhang, Xiangyu and Ren, Shaoqing and Sun, Jian},
  booktitle={Proceedings of the IEEE conference on computer vision and pattern recognition},
  pages={770--778},
  year={2016}
}

@article{simonyan2014very,
  title={Very deep convolutional networks for large-scale image recognition},
  author={Simonyan, Karen and Zisserman, Andrew},
  journal={arXiv preprint arXiv:1409.1556},
  year={2014}
}

@article{vaswani2017attention,
  title={Attention is all you need},
  author={Vaswani, Ashish and Shazeer, Noam and Parmar, Niki and Uszkoreit, Jakob and Jones, Llion and Gomez, Aidan N and Kaiser, {\L}ukasz and Polosukhin, Illia},
  journal={Advances in neural information processing systems},
  volume={30},
  year={2017}
}

@inproceedings{hu2018cvm,
  title={Cvm-net: Cross-view matching network for image-based ground-to-aerial geo-localization},
  author={Hu, Sixing and Feng, Mengdan and Nguyen, Rang MH and Lee, Gim Hee},
  booktitle={Proceedings of the IEEE Conference on Computer Vision and Pattern Recognition},
  pages={7258--7267},
  year={2018}
}

@inproceedings{workman2015wide,
  title={Wide-area image geolocalization with aerial reference imagery},
  author={Workman, Scott and Souvenir, Richard and Jacobs, Nathan},
  booktitle={Proceedings of the IEEE International Conference on Computer Vision},
  pages={3961--3969},
  year={2015}
}

@article{dosovitskiy2020image,
  title={An image is worth 16x16 words: Transformers for image recognition at scale},
  author={Dosovitskiy, Alexey and Beyer, Lucas and Kolesnikov, Alexander and Weissenborn, Dirk and Zhai, Xiaohua and Unterthiner, Thomas and Dehghani, Mostafa and Minderer, Matthias and Heigold, Georg and Gelly, Sylvain and others},
  journal={arXiv preprint arXiv:2010.11929},
  year={2020}
}

@inproceedings{liu2021swin,
  title={Swin transformer: Hierarchical vision transformer using shifted windows},
  author={Liu, Ze and Lin, Yutong and Cao, Yue and Hu, Han and Wei, Yixuan and Zhang, Zheng and Lin, Stephen and Guo, Baining},
  booktitle={Proceedings of the IEEE/CVF International Conference on Computer Vision},
  pages={10012--10022},
  year={2021}
}

@inproceedings{bertasius2021space,
  title={Is space-time attention all you need for video understanding?},
  author={Bertasius, Gedas and Wang, Heng and Torresani, Lorenzo},
  booktitle={ICML},
  volume={2},
  number={3},
  pages={4},
  year={2021}
}

@inproceedings{liu2019lending,
  title={Lending orientation to neural networks for cross-view geo-localization},
  author={Liu, Liu and Li, Hongdong},
  booktitle={Proceedings of the IEEE/CVF Conference on Computer Vision and Pattern Recognition},
  pages={5624--5633},
  year={2019}
}

@article{regmi2019cross,
  title={Cross-view image synthesis using geometry-guided conditional gans},
  author={Regmi, Krishna and Borji, Ali},
  journal={Computer Vision and Image Understanding},
  volume={187},
  pages={102788},
  year={2019},
  publisher={Elsevier}
}

@inproceedings{regmi2019bridging,
  title={Bridging the domain gap for ground-to-aerial image matching},
  author={Regmi, Krishna and Shah, Mubarak},
  booktitle={Proceedings of the IEEE/CVF International Conference on Computer Vision},
  pages={470--479},
  year={2019}
}

@inproceedings{shi2020looking,
  title={Where am i looking at? joint location and orientation estimation by cross-view matching},
  author={Shi, Yujiao and Yu, Xin and Campbell, Dylan and Li, Hongdong},
  booktitle={Proceedings of the IEEE/CVF Conference on Computer Vision and Pattern Recognition},
  pages={4064--4072},
  year={2020}
}

@inproceedings{toker2021coming,
  title={Coming down to earth: Satellite-to-street view synthesis for geo-localization},
  author={Toker, Aysim and Zhou, Qunjie and Maximov, Maxim and Leal-Taix{\'e}, Laura},
  booktitle={Proceedings of the IEEE/CVF Conference on Computer Vision and Pattern Recognition},
  pages={6488--6497},
  year={2021}
}

@inproceedings{yu2020bdd100k,
  title={Bdd100k: A diverse driving dataset for heterogeneous multitask learning},
  author={Yu, Fisher and Chen, Haofeng and Wang, Xin and Xian, Wenqi and Chen, Yingying and Liu, Fangchen and Madhavan, Vashisht and Darrell, Trevor},
  booktitle={Proceedings of the IEEE/CVF conference on computer vision and pattern recognition},
  pages={2636--2645},
  year={2020}
}

@inproceedings{regmi2021video,
  title={Video geo-localization employing geo-temporal feature learning and gps trajectory smoothing},
  author={Regmi, Krishna and Shah, Mubarak},
  booktitle={Proceedings of the IEEE/CVF International Conference on Computer Vision},
  pages={12126--12135},
  year={2021}
}

@inproceedings{vyas2022gama,
  title={GAMa: Cross-view Video Geo-localization},
  author={Vyas, Shruti and Chen, Chen and Shah, Mubarak},
  booktitle={European Conference on Computer Vision},
  pages={440--456},
  year={2022},
  organization={Springer}
}

@inproceedings{zhu2022transgeo,
  title={TransGeo: Transformer Is All You Need for Cross-view Image Geo-localization},
  author={Zhu, Sijie and Shah, Mubarak and Chen, Chen},
  booktitle={Proceedings of the IEEE/CVF Conference on Computer Vision and Pattern Recognition},
  pages={1162--1171},
  year={2022}
}

@article{yang2021cross,
  title={Cross-view geo-localization with layer-to-layer transformer},
  author={Yang, Hongji and Lu, Xiufan and Zhu, Yingying},
  journal={Advances in Neural Information Processing Systems},
  volume={34},
  pages={29009--29020},
  year={2021}
}

@inproceedings{zhu2021vigor,
  title={Vigor: Cross-view image geo-localization beyond one-to-one retrieval},
  author={Zhu, Sijie and Yang, Taojiannan and Chen, Chen},
  booktitle={Proceedings of the IEEE/CVF Conference on Computer Vision and Pattern Recognition},
  pages={3640--3649},
  year={2021}
}

@article{paszke2019pytorch,
  title={Pytorch: An imperative style, high-performance deep learning library},
  author={Paszke, Adam and Gross, Sam and Massa, Francisco and Lerer, Adam and Bradbury, James and Chanan, Gregory and Killeen, Trevor and Lin, Zeming and Gimelshein, Natalia and Antiga, Luca and others},
  journal={Advances in neural information processing systems},
  volume={32},
  year={2019}
}

@inproceedings{weyand2016planet,
  title={Planet-photo geolocation with convolutional neural networks},
  author={Weyand, Tobias and Kostrikov, Ilya and Philbin, James},
  booktitle={European Conference on Computer Vision},
  pages={37--55},
  year={2016},
  organization={Springer}
}

@inproceedings{pramanick2022world,
  title={Where in the World is this Image? Transformer-based Geo-localization in the Wild},
  author={Pramanick, Shraman and Nowara, Ewa M and Gleason, Joshua and Castillo, Carlos D and Chellappa, Rama},
  booktitle={European Conference on Computer Vision},
  pages={196--215},
  year={2022},
  organization={Springer}
}

@inproceedings{muller2018geolocation,
  title={Geolocation estimation of photos using a hierarchical model and scene classification},
  author={Muller-Budack, Eric and Pustu-Iren, Kader and Ewerth, Ralph},
  booktitle={Proceedings of the European Conference on Computer Vision (ECCV)},
  pages={563--579},
  year={2018}
}

@inproceedings{clark2023we,
  title={Where We Are and What We're Looking At: Query Based Worldwide Image Geo-Localization Using Hierarchies and Scenes},
  author={Clark, Brandon and Kerrigan, Alec and Kulkarni, Parth Parag and Cepeda, Vicente Vivanco and Shah, Mubarak},
  booktitle={Proceedings of the IEEE/CVF Conference on Computer Vision and Pattern Recognition},
  pages={23182--23190},
  year={2023}
}

@inproceedings{seo2018cplanet,
  title={Cplanet: Enhancing image geolocalization by combinatorial partitioning of maps},
  author={Seo, Paul Hongsuck and Weyand, Tobias and Sim, Jack and Han, Bohyung},
  booktitle={Proceedings of the European Conference on Computer Vision (ECCV)},
  pages={536--551},
  year={2018}
}

@inproceedings{warburg2020mapillary,
  title={Mapillary street-level sequences: A dataset for lifelong place recognition},
  author={Warburg, Frederik and Hauberg, Soren and Lopez-Antequera, Manuel and Gargallo, Pau and Kuang, Yubin and Civera, Javier},
  booktitle={Proceedings of the IEEE/CVF conference on computer vision and pattern recognition},
  pages={2626--2635},
  year={2020}
}

@ARTICLE{6165309,
  author={Ji, Shuiwang and Xu, Wei and Yang, Ming and Yu, Kai},
  journal={IEEE Transactions on Pattern Analysis and Machine Intelligence}, 
  title={3D Convolutional Neural Networks for Human Action Recognition}, 
  year={2013},
  volume={35},
  number={1},
  pages={221-231},
  doi={10.1109/TPAMI.2012.59}}

@inproceedings{hays2008im2gps,
  title={Im2gps: estimating geographic information from a single image},
  author={Hays, James and Efros, Alexei A},
  booktitle={2008 ieee conference on computer vision and pattern recognition},
  pages={1--8},
  year={2008},
  organization={IEEE}
}

@inproceedings{vo2017revisiting,
  title={Revisiting im2gps in the deep learning era},
  author={Vo, Nam and Jacobs, Nathan and Hays, James},
  booktitle={Proceedings of the IEEE international conference on computer vision},
  pages={2621--2630},
  year={2017}
}

@article{larson2017benchmarking,
  title={The benchmarking initiative for multimedia evaluation: MediaEval 2016},
  author={Larson, Martha and Soleymani, Mohammad and Gravier, Guillaume and Ionescu, Bogdan and Jones, Gareth JF},
  journal={IEEE MultiMedia},
  volume={24},
  number={1},
  pages={93--96},
  year={2017},
  publisher={IEEE}
}

@article{tong2022videomae,
  title={Videomae: Masked autoencoders are data-efficient learners for self-supervised video pre-training},
  author={Tong, Zhan and Song, Yibing and Wang, Jue and Wang, Limin},
  journal={Advances in neural information processing systems},
  volume={35},
  pages={10078--10093},
  year={2022}
}

@article{zhou2017places,
  title={Places: A 10 million image database for scene recognition},
  author={Zhou, Bolei and Lapedriza, Agata and Khosla, Aditya and Oliva, Aude and Torralba, Antonio},
  journal={IEEE transactions on pattern analysis and machine intelligence},
  volume={40},
  number={6},
  pages={1452--1464},
  year={2017},
  publisher={IEEE}
}

@article{kingma2014adam,
  title={Adam: A method for stochastic optimization},
  author={Kingma, Diederik P and Ba, Jimmy},
  journal={arXiv preprint arXiv:1412.6980},
  year={2014}
}

@inproceedings{radford2021learning,
  title={Learning transferable visual models from natural language supervision},
  author={Radford, Alec and Kim, Jong Wook and Hallacy, Chris and Ramesh, Aditya and Goh, Gabriel and Agarwal, Sandhini and Sastry, Girish and Askell, Amanda and Mishkin, Pamela and Clark, Jack and others},
  booktitle={International conference on machine learning},
  pages={8748--8763},
  year={2021},
  organization={PMLR}
}

@article{vivanco2024geoclip,
  title={GeoCLIP: Clip-Inspired Alignment between Locations and Images for Effective Worldwide Geo-localization},
  author={Vivanco Cepeda, Vicente and Nayak, Gaurav Kumar and Shah, Mubarak},
  journal={Advances in Neural Information Processing Systems},
  volume={36},
  year={2024}
}

@article{radford2019language,
  title={Language models are unsupervised multitask learners},
  author={Radford, Alec and Wu, Jeffrey and Child, Rewon and Luan, David and Amodei, Dario and Sutskever, Ilya and others},
  journal={OpenAI blog},
  volume={1},
  number={8},
  pages={9},
  year={2019}
}

@inproceedings{vaca2012city,
  title={City scale geo-spatial trajectory estimation of a moving camera},
  author={Vaca-Castano, Gonzalo and Zamir, Amir Roshan and Shah, Mubarak},
  booktitle={2012 IEEE Conference on Computer Vision and Pattern Recognition},
  pages={1186--1193},
  year={2012},
  organization={IEEE}
}

@inproceedings{kulkarni2024cityguessr,
  title={CityGuessr: City-Level Video Geo-Localization on a Global Scale},
  author={Kulkarni, Parth Parag and Nayak, Gaurav Kumar and Shah, Mubarak},
  booktitle={European Conference on Computer Vision},
  pages={293--311},
  year={2024},
  organization={Springer}
}

@article{oquab2023dinov2,
  title={Dinov2: Learning robust visual features without supervision},
  author={Oquab, Maxime and Darcet, Timoth{\'e}e and Moutakanni, Th{\'e}o and Vo, Huy and Szafraniec, Marc and Khalidov, Vasil and Fernandez, Pierre and Haziza, Daniel and Massa, Francisco and El-Nouby, Alaaeldin and others},
  journal={arXiv preprint arXiv:2304.07193},
  year={2023}
}

@inproceedings{haas2024pigeon,
  title={Pigeon: Predicting image geolocations},
  author={Haas, Lukas and Skreta, Michal and Alberti, Silas and Finn, Chelsea},
  booktitle={Proceedings of the IEEE/CVF Conference on Computer Vision and Pattern Recognition},
  pages={12893--12902},
  year={2024}
}

@article{pillai2024garet,
  title={GAReT: Cross-view Video Geolocalization with Adapters and Auto-Regressive Transformers},
  author={Pillai, Manu S and Rizve, Mamshad Nayeem and Shah, Mubarak},
  journal={arXiv preprint arXiv:2408.02840},
  year={2024}
}

@article{vsavrivc2019equal,
  title={The equal earth map projection},
  author={{\v{S}}avri{\v{c}}, Bojan and Patterson, Tom and Jenny, Bernhard},
  journal={International Journal of Geographical Information Science},
  volume={33},
  number={3},
  pages={454--465},
  year={2019},
  publisher={Taylor \& Francis}
}

@techreport{eiter1994computing,
  author      = {Thomas Eiter and Heikki Mannila},
  title       = {Computing Discrete Fr{\'e}chet Distance},
  year        = {1994},
  institution = {Christian Doppler Laboratory for Expert Systems},
  number      = {CD-TR 94/64},
  type        = {Technical Report}
}

@article{tancik2020fourier,
  title={Fourier features let networks learn high frequency functions in low dimensional domains},
  author={Tancik, Matthew and Srinivasan, Pratul and Mildenhall, Ben and Fridovich-Keil, Sara and Raghavan, Nithin and Singhal, Utkarsh and Ramamoorthi, Ravi and Barron, Jonathan and Ng, Ren},
  journal={Advances in neural information processing systems},
  volume={33},
  pages={7537--7547},
  year={2020}
}

@inproceedings{gaharwar2023xi,
  title={Xi-Net: Transformer based Seismic Waveform Reconstructor},
  author={Gaharwar, Anshuman and Kulkarni, Parth Parag and Dickey, Joshua and Shah, Mubarak},
  booktitle={2023 IEEE International Conference on Image Processing (ICIP)},
  pages={2725--2729},
  year={2023},
  organization={IEEE}
}

@article{mi2024congeo,
  title={ConGeo: Robust Cross-view Geo-localization across Ground View Variations},
  author={Mi, Li and Xu, Chang and Castillo-Navarro, Javiera and Montariol, Syrielle and Yang, Wen and Bosselut, Antoine and Tuia, Devis},
  journal={arXiv preprint arXiv:2403.13965},
  year={2024}
}

@inproceedings{deuser2023sample4geo,
  title={Sample4geo: Hard negative sampling for cross-view geo-localisation},
  author={Deuser, Fabian and Habel, Konrad and Oswald, Norbert},
  booktitle={Proceedings of the IEEE/CVF International Conference on Computer Vision},
  pages={16847--16856},
  year={2023}
}

@article{lewis2019bart,
  title={Bart: Denoising sequence-to-sequence pre-training for natural language generation, translation, and comprehension},
  author={Lewis, M},
  journal={arXiv preprint arXiv:1910.13461},
  year={2019}
}

@article{wang2020fairseq,
  title={Fairseq S2T: Fast speech-to-text modeling with fairseq},
  author={Wang, Changhan and Tang, Yun and Ma, Xutai and Wu, Anne and Popuri, Sravya and Okhonko, Dmytro and Pino, Juan},
  journal={arXiv preprint arXiv:2010.05171},
  year={2020}
}

@article{misra2019mish,
  title={Mish: A self regularized non-monotonic activation function},
  author={Misra, Diganta},
  journal={arXiv preprint arXiv:1908.08681},
  year={2019}
}

@inproceedings{rezatofighi2019generalized,
  title={Generalized intersection over union: A metric and a loss for bounding box regression},
  author={Rezatofighi, Hamid and Tsoi, Nathan and Gwak, JunYoung and Sadeghian, Amir and Reid, Ian and Savarese, Silvio},
  booktitle={Proceedings of the IEEE/CVF conference on computer vision and pattern recognition},
  pages={658--666},
  year={2019}
}

@article{pearson1901liii,
  title={LIII. On lines and planes of closest fit to systems of points in space},
  author={Pearson, Karl},
  journal={The London, Edinburgh, and Dublin philosophical magazine and journal of science},
  volume={2},
  number={11},
  pages={559--572},
  year={1901},
  publisher={Taylor \& Francis}
}

@inproceedings{zhai2023sigmoid,
  title={Sigmoid loss for language image pre-training},
  author={Zhai, Xiaohua and Mustafa, Basil and Kolesnikov, Alexander and Beyer, Lucas},
  booktitle={Proceedings of the IEEE/CVF international conference on computer vision},
  pages={11975--11986},
  year={2023}
}

@article{bai2025qwen2,
  title={Qwen2. 5-vl technical report},
  author={Bai, Shuai and Chen, Keqin and Liu, Xuejing and Wang, Jialin and Ge, Wenbin and Song, Sibo and Dang, Kai and Wang, Peng and Wang, Shijie and Tang, Jun and others},
  journal={arXiv preprint arXiv:2502.13923},
  year={2025}
}

@inproceedings{li2024georeasoner,
  title={Georeasoner: Geo-localization with reasoning in street views using a large vision-language model},
  author={Li, Ling and Ye, Yu and Jiang, Bingchuan and Zeng, Wei},
  booktitle={Forty-first International Conference on Machine Learning},
  year={2024}
}

@article{campos2025gaea,
  title={Gaea: A geolocation aware conversational model},
  author={Campos, Ron and Vayani, Ashmal and Kulkarni, Parth Parag and Gupta, Rohit and Dutta, Aritra and Shah, Mubarak},
  journal={arXiv preprint arXiv:2503.16423},
  year={2025}
}

@inproceedings{shatwell2025gt,
  title={GT-Loc: Unifying When and Where in Images Through a Joint Embedding Space},
  author={Shatwell, David G and Dave, Ishan Rajendrakumar and Swetha, Sirnam and Shah, Mubarak},
  booktitle={Proceedings of the IEEE/CVF International Conference on Computer Vision},
  pages={1--11},
  year={2025}
}

@article{lewis2020retrieval,
  title={Retrieval-augmented generation for knowledge-intensive nlp tasks},
  author={Lewis, Patrick and Perez, Ethan and Piktus, Aleksandra and Petroni, Fabio and Karpukhin, Vladimir and Goyal, Naman and K{\"u}ttler, Heinrich and Lewis, Mike and Yih, Wen-tau and Rockt{\"a}schel, Tim and others},
  journal={Advances in neural information processing systems},
  volume={33},
  pages={9459--9474},
  year={2020}
}

@article{jia2024g3,
  title={G3: an effective and adaptive framework for worldwide geolocalization using large multi-modality models},
  author={Jia, Pengyue and Liu, Yiding and Li, Xiaopeng and Zhao, Xiangyu and Wang, Yuhao and Du, Yantong and Han, Xiao and Wei, Xuetao and Wang, Shuaiqiang and Yin, Dawei},
  journal={Advances in Neural Information Processing Systems},
  volume={37},
  pages={53198--53221},
  year={2024}
}

@article{jia2025georanker,
  title={GeoRanker: Distance-Aware Ranking for Worldwide Image Geolocalization},
  author={Jia, Pengyue and Park, Seongheon and Gao, Song and Zhao, Xiangyu and Li, Yixuan},
  journal={arXiv preprint arXiv:2505.13731},
  year={2025}
}

@inproceedings{astruc2024openstreetview,
  title={Openstreetview-5m: The many roads to global visual geolocation},
  author={Astruc, Guillaume and Dufour, Nicolas and Siglidis, Ioannis and Aronssohn, Constantin and Bouia, Nacim and Fu, Stephanie and Loiseau, Romain and Nguyen, Van Nguyen and Raude, Charles and Vincent, Elliot and others},
  booktitle={Proceedings of the IEEE/CVF Conference on Computer Vision and Pattern Recognition},
  pages={21967--21977},
  year={2024}
}

@article{haas2023learning,
  title={Learning generalized zero-shot learners for open-domain image geolocalization},
  author={Haas, Lukas and Alberti, Silas and Skreta, Michal},
  journal={arXiv preprint arXiv:2302.00275},
  year={2023}
}

@article{lecun1989handwritten,
  title={Handwritten digit recognition with a back-propagation network},
  author={LeCun, Yann and Boser, Bernhard and Denker, John and Henderson, Donnie and Howard, Richard and Hubbard, Wayne and Jackel, Lawrence},
  journal={Advances in neural information processing systems},
  volume={2},
  year={1989}
}
}

\clearpage
\setcounter{page}{1}
\maketitlesupplementary

In this supplementary material, we provide additional insights on our method. We organize our supplementary into the following sections:
\begin{enumerate}[label=\Alph*.]  
    \item Training Details
    \item Additional Ablations
    \begin{enumerate}[label=B.\arabic*.]
        \item Ablation on $\alpha$ and $\beta$ in Training Phase
        \item Ablation on MLP layers in Image Encoder
        \item Ablation on $\sigma$ values for RFF in Location Encoder
        \item Ablation on Training sequence length
    \end{enumerate}
    \item Analysis of Predictions by Video Length
    \item Reference Gallery
    \begin{enumerate}[label=D.\arabic*.]
        \item Uniform Grid Gallery Construction
        \item Evaluation with Val-set Gallery Retrieval
    \end{enumerate}
    \item Video and Trajectory Metrics
    \begin{enumerate}[label=E.\arabic*.]  
        \item Video Distance Threshold Accuracy 
        \item Discrete Frechet Distance (DFD)
        \item Mean Range Difference (MRD)
    \end{enumerate}
    \item Training of Baselines/SoTA models
    \begin{enumerate}[label=F.\arabic*.]
        \item Zero Shot
        \item Classification-based SoTA models
        \item Classification-based Foundation Baselines
        \item Retrieval-based SoTA model
    \end{enumerate}
    \item Visualizations in Feature Space
    \item Unified Model
    \item Computational Cost
    \item Qualitative Results on GaMa
    \item City-level Geolocalization Performance Analysis on CityGuessr68k
    \item Limitations and Failure Analysis
    \item Examples of Localizations 
    
\end{enumerate}

\appendix

\section{Training Details}
\label{sec:td}
Our model is implemented in PyTorch\cite{paszke2019pytorch}. The input video consists of 16 frames for training in both phases, resized to 224x224. The model is trained on one node of an NVIDIA RTX A6000 GPU. Both DINOv2\cite{oquab2023dinov2} and CLIP\cite{radford2021learning} versions used, have a ViT-L/14\cite{dosovitskiy2020image} backbone. Phase I is trained for 600 epochs while Phase II is trained for 100. For both phases batch size is 128 sequences. We use Adam\cite{kingma2014adam} optimizer with a StepLR scheduler with an learning rate decay of 0.99 for Phase I, and 0.95 for Phase II, with 1000 steps of warmup. For Phase I, learning rate is 5e-5 while for Phase II it is 1e-4. The TempGeo module has 2 transformer\cite{vaswani2017attention} encoder layers, while the GeoRefiner module has 1 transformer encoder layer and 2 transformer decoder layers. For Phase II noise addition, entire sequence is collapsed to a single point with a 10\% chance. The rest of the time a random jitter is added with a random sign to both GPS coordinates individually, sampled from a uniform distribution between 0.001 and 0.02. The entire sequence is then shifted with an added noise sampled from a uniform distribution between -0.2 and 0.2. For weighted Hinge loss, $\alpha = 10$, $\beta = 1$. 

For training on GaMa\cite{vyas2022gama} and CityGuessr68k\cite{kulkarni2024cityguessr} some of the settings are slightly different. Both GaMa split of the BDD100k\cite{yu2020bdd100k} dataset and CityGuessr68k are much larger datasets than Mapillary(MSLS)\cite{warburg2020mapillary}. However, as described in Section \ref{sec:data}, GaMa lacks geographical coverage, while CityGuessr68k lacks GPS annotation and is relatively coarse-grained. In GaMa, each video has approximately 38 frames, while number of frames for videos in CityGuessr68k is approximately 100. For consistency with Mapillary(MSLS) which has atleast 16 frames per video, 16 frames were sampled from both datasets.

For both model versions, the training was conducted for 100 epochs with a learning rate of 5e-5. Similar to the Mapillary(MSLS) model version, Adam\cite{kingma2014adam} optimizer was used with a StepLR scheduler, learning rate decay of 0.95 and 1000 steps of warmup. The images were resized to 224x224 and training was conducted on a single node of an NVIDIA RTX A6000 GPU. An important observation was that the model version trained on CityGuessr68k converged a lot quicker than the models trained on other two datasets. Note that the training details change only for Phase I, as Phase II input dimension is dataset invariant.

\section{Additional Ablations}
\label{sec:add-abl}
In this section, we include additional ablations for the choice of $\alpha$ and $\beta$ hyperparameters, number of MLP layers in image encoder, $\sigma$ values for RFF in location encoder and training sequence length.

\subsection{Ablation on $\alpha$ and $\beta$ in Training Phase}
\label{sec:abl-alphabeta}
Phase II of our training focuses on alignment of the refined GPS features to the ground truth GPS features. As a result, we opt for using a weighted Hinge loss for training our model. As discussed in Main paper Section \ref{sec:loss}, we use hyperparameters $\alpha$ and $\beta$ to denote the weights of negative pairs(upper triangular and lower triangular elements of the similarity matrix) and weights of positive pairs(diagonal elements of the similarity matrix) respectively. Table \ref{tab:alphabeta} shows that $\alpha=10$ and $\beta = 1$ gives the best results at most distance thresholds as well as in terms of median distance error and trajectory metrics DFD and MRD.

In a similarity matrix of size N, there are $N$ diagonal elements and $N^2-N$ off-diagonal 
elements. Thus the number of diagonal elements scale linearly while the off-diagonal elements scale quadratically. This implies that the weight of off-diagonal elements is a lot more than the diagonal elements which is undesirable for feature alignment. This scaling is exacerbated in the frame-wise component of the loss where $N = batch\_size \times number\_of\_frames$. The fixed weights $\alpha$ and $\beta$ mitigates this scaling, and assists the model to align diagonal elements better. The values of $\alpha=10$ and $\beta=1$ are chosen such that the negative and positive terms of the loss have the same order of magnitude. 

\begin{table}[!h]
\caption{Ablation for values of $\alpha$ and $\beta$}
\resizebox{0.48\textwidth}{!}{
\begin{tabular}{c|cccc|c|c|c}
\hline
\multirow{3}{*}{$\alpha$,$\beta$} & \multicolumn{4}{c|}{\textbf{Distance ($a_r$ {[}\%{]} @ km) $\uparrow$}}               & \multirow{3}{*}{\begin{tabular}[c]{@{}c@{}}\textbf{Median}\\ \textbf{Distance}\\ \textbf{Error(km)} $\downarrow$\end{tabular}} & \multirow{3}{*}{\textbf{DFD} $\downarrow$} & \multirow{3}{*}{\textbf{MRD} $\downarrow$}\\
                            & \textbf{Sub-street}     & \textbf{Street}        & \textbf{Locality}      & \textbf{City}          &                                                                                        &                      \\
                            & \textbf{0.5 km (500 m)} & \textbf{1 km}          & \textbf{5 km}          & \textbf{25 km}         &                                                                                        &                      \\ \hline
1,1                         & \textbf{21.5}           &     \textbf{41.3}      & 75.5 & 97.4 & 1.36                         &     5.21 &             1.41                                                            \\
10,1                        & \textbf{21.5}  & 41.0 &     \textbf{76.7}      & \textbf{97.9} & \textbf{1.35}                                   & \textbf{3.87}                                       &   \textbf{1.07}                   \\
100,1                       &     21.4       &    40.9       & 75.4          & 97.2          & 1.38                                                 &      7.63                             &    1.57                  \\ \hline
\end{tabular}
}
\label{tab:alphabeta}
\end{table}

\subsection{Ablation on MLP layers in Image Encoder}
\label{sec:abl-mlp}
As we can see in Main paper Fig. \ref{fig:arch}, the frame encoder consists of two backbone encoders, DINOv2\cite{oquab2023dinov2} and CLIP\cite{radford2021learning}, the TempGeo module and an MLP. Frames of a video are passed through both DINOv2 and CLIP backbones, and both sets of features are concatenated. The combined features for all frames of a video are input into the TempGeo module that aligns those frames with each other. Finally the output of TempGeo is input to the MLP which gives the final embeddings corresponding to every video frame. The main purpose of the MLP is to accommodate the dimension of the frame embeddings, such that they become the same size as the GPS embeddings which are output from the location encoder. The output dimensions of vectors from DINOv2(ViT-L/14\cite{dosovitskiy2020image}) and CLIP(ViT-L/14) are 1024 and 768 respectively. A 1792-dimensional vector is formed by concatenating the two. As TempGeo is purely a transformer architecture, the input and output are of the same size. The desired final frame embedding dimension is 512. Thus an MLP is employed for converting the 1792-dimensional output of TempGeo to a 512-dimensional frame embedding. Our VidTAG model consists of a 3-layer MLP with Mish\cite{misra2019mish} activations in between. The Vector dimension through the MLP goes from 1792 to 1024(layer 1), from 1024 to 768(layer 2) and finally from 768 to 512, making this a gentle transition. 

For analysis, we trained variations of our model with different number of MLP layers. These model variations were trained on 100 epochs and a learning rate decay of 0.95 for efficiency. Rest all settings were kept equivalent to the main model. As shown in Table \ref{tab:mlp}, the choice of 3 layers is the most adequate, as further addition of more layers results in diminishing returns or a decrease in performance, while increasing the number of trainable parameters. We see that at the locality level, the model with 4 MLP layers performs the best. However, for the finer thresholds, and on median distance error, the 3-layer version performs the best, which justifies the setting in our model as those metrics are of more importance for the task at hand. 

\begin{table}[!h]
\caption{Ablation on Number of layers in MLP in Image Encoder}
\resizebox{0.48\textwidth}{!}{
\begin{tabular}{c|cccc|c|c}
\hline
\multirow{3}{*}{\begin{tabular}[c]{@{}c@{}}\textbf{No. of}\\ \textbf{MLP Layers}\end{tabular}} & \multicolumn{4}{c|}{\textbf{Distance ($a_r$ {[}\%{]} @ km) $\uparrow$}}                                                                                                                                                                                           & \multirow{2}{*}{\begin{tabular}[c]{@{}c@{}}\textbf{Median}\\ \textbf{Distance}\\ \textbf{Error(km)} $\downarrow$\end{tabular}} & \multirow{3}{*}{\begin{tabular}[c]{@{}c@{}}\textbf{Trainable}\\ \textbf{Parameters}\end{tabular}}\\
                                   & \begin{tabular}[c]{@{}c@{}}\textbf{Sub-street}\\ \textbf{0.5 km (500 m)}\end{tabular} & \begin{tabular}[c]{@{}c@{}}\textbf{Street}\\ \textbf{1 km}\end{tabular} & \begin{tabular}[c]{@{}c@{}}\textbf{Locality}\\ \textbf{5 km}\end{tabular} & \begin{tabular}[c]{@{}c@{}}\textbf{City}\\ \textbf{25 km}\end{tabular} &                                                                                        \\ \hline
2                                  &      12.5                                                              &                                 32.6                      &      75.4                                                 &                                     98.5                 &                1.71                                             &        55.7M               \\
3                                  &   13.5                                                      &                          \textbf{33.1}                             &        \textbf{77.4}                                             &                     98.7                                 &              \textbf{1.66}   &          56.3M                              \\
4                                  &        \textbf{13.8}                                                             &                                     31.9                 &           75.6                                              &                         98.1                             &           1.72                                                 &      57.4M                       \\
5                                  &         12.8                                                        &                             30.6                      &               76.2                                     &                                        \textbf{98.8}           &           1.82                                           &      58M                      \\ \hline
\end{tabular}
}
\label{tab:mlp}
\end{table}
\subsection{Ablation on $\sigma$ values for RFF in Location Encoder}
\label{sec:abl-rff}
The location encoder is used to obtain viable embeddings from GPS coordinates so that they can be aligned with frame embeddings in the same feature space. The concept of the location encoder was first introduced in GeoCLIP\cite{vivanco2024geoclip}. The location encoder computes Random Fourier Features(RFF)\cite{tancik2020fourier} of the Equal Earth Projections(EEP)\cite{vsavrivc2019equal} of GPS coordinates at different frequencies. These RFF values are passed through individual MLPs and then added together to output the final GPS embedding.  $\sigma$ values play an important role in determining the range of these frequencies for computing RFF. GeoCLIP inferred that larger $\sigma$ values are preferable for finer scales while smaller $\sigma$ values perform better at more coarse-grained level. Having said that, determining the exact combination of $\sigma$ values that work best for a particular model is not straightforward and requires extensive amount of experimentation. In Table \ref{tab:rff}, we show the performance of our model trained on a variety of combinations of 3 frequencies determined by the shown $\sigma$ values ($\sigma_0$, $\sigma_1$, $\sigma_2$). We vary $\sigma_0$ from $2^0$ to $2^2$, $\sigma_1$ from $2^3$ to $2^5$ and $\sigma_2$ from $2^7$ to $2^9$. We observe that the combination of [$\sigma_0 = 2^0$, $\sigma_1 = 2^3$, $\sigma_2 = 2^8$] works the best and provides the model with the highest performance across most of the metrics.
\begin{table}[!h]
\caption{Ablation on $\sigma$ values for RFF in Location Encoder}
\resizebox{0.5\textwidth}{!}{
\begin{tabular}{ccc|cccc|c}
\hline
\multicolumn{3}{c|}{\textbf{$\sigma$ values}}                                                   & \multicolumn{4}{c|}{\textbf{Distance ($a_r$ {[}\%{]} @ km) $\uparrow$}}                                                                                                                                                                                           & \multirow{2}{*}{\begin{tabular}[c]{@{}c@{}}\textbf{Median}\\ \textbf{Distance}\\ \textbf{Error(km)} $\downarrow$\end{tabular}} \\
$\sigma_0$             & \multicolumn{1}{l}{$\sigma_1$} & \multicolumn{1}{l|}{$\sigma_2$} & \begin{tabular}[c]{@{}c@{}}\textbf{Sub-street}\\ \textbf{0.5 km (500 m)}\end{tabular} & \begin{tabular}[c]{@{}c@{}}\textbf{Street}\\ \textbf{1 km}\end{tabular} & \begin{tabular}[c]{@{}c@{}}\textbf{Locality}\\ \textbf{5 km}\end{tabular} & \begin{tabular}[c]{@{}c@{}}\textbf{City}\\ \textbf{25 km}\end{tabular} &                                                                                        \\ \hline
$2^0$ & $2^3$         & $2^8$          &              \textbf{13.5}                                                       &                          \textbf{33.1}                             &        \textbf{77.4}                                             &                     98.7                                 &              \textbf{1.66}                                                                         \\
$2^0$ & $2^4$         & $2^7$          &       \textbf{13.5}                                                            &                                 32.5                      &             \textbf{77.4}                                            &                  98.5                                   &            1.71                                                                            \\
$2^0$ & $2^4$         & $2^8$          &       13.0                                                          &                                        32.7           &          75.4                                           &     98.7                                             &              1.68                                                                      \\
$2^0$ & $2^4$         & $2^9$          &        13.3                                                             &                             32.3                        &              77.3                                           &                        98.8                           &       1.71                                                                               \\
$2^0$ & $2^5$         & $2^8$          &          13.1                                                          &                                     31.5                 &                     76.6                                    &                               98.8                     &             1.78                                                                           \\
$2^1$ & $2^4$         & $2^8$          &       12.8                                                            &                                  31.3                  &          75.6                                             &                                      98.5               &           1.73                                                                           \\
$2^2$ & $2^4$         & $2^8$          &          13.3                                                         &                                   33.0              &                75.8                                       &                                \textbf{98.8}                &           1.81                                                                          \\ \hline
\end{tabular}
}
\label{tab:rff}
\end{table}

\subsection{Ablation on Training sequence length}
\label{sec:abl-tsl}
As stated perviously, every video in our training split of Mapillary(MSLS) has atleast 16 frames. Length of training sequences does have an impact on the performance of the model. Generally, longer training sequences result in better performing models which can happen due to a multitude of reasons. Longer training sequences imply more information per sequence that the model sees, thus generating better features. Also, as validation sequences are longer and of arbitrary length, getting the model used to longer sequences is beneficial. Finally, the TempGeo module gets more frames as reference, for feature alignment. For efficiency, these ablations were performed on similar settings to the ones described in Section \ref{sec:abl-mlp}. Table \ref{tab:tsl} shows similar behavior as we have anticipated above. We see a steady increase in the models performance on almost all thresholds as we increase the length of training sequences from 4 to 8 to 16 frames. These frames are sampled from the larger training videos, varying just the number of frames sampled.  We see that the accuracy of the model at 1km consistently increases from 26.6\% for 4 frames to 31.3\% for 8 frames to finally 33.1\% for 16 frames. The median error is also reduced from 2.13 km to 1.66 km. Thus we use sequences of 16 frames to train our final model.  
\begin{table}[!h]
\caption{Ablation on Training Sequence Length}
\resizebox{0.48\textwidth}{!}{
\begin{tabular}{c|cccc|c}
\hline
\multirow{3}{*}{\begin{tabular}[c]{@{}c@{}}\textbf{Training Sequence Length}\\ \textbf{(No. of frames)}\end{tabular}} & \multicolumn{4}{c|}{\textbf{Distance ($a_r$ {[}\%{]} @ km) $\uparrow$}}                                                                                                                                                                                  & \multirow{2}{*}{\begin{tabular}[c]{@{}c@{}}\textbf{Median}\\ \textbf{Distance}\\ \textbf{Error(km)} $\downarrow$\end{tabular}} \\
                                                                                                    & \begin{tabular}[c]{@{}c@{}}\textbf{Sub-street}\\ \textbf{0.5 km (500 m)}\end{tabular} & \begin{tabular}[c]{@{}c@{}}\textbf{Street}\\ \textbf{1 km}\end{tabular} & \begin{tabular}[c]{@{}c@{}}\textbf{Locality}\\ \textbf{5 km}\end{tabular} & \begin{tabular}[c]{@{}c@{}}\textbf{City}\\ \textbf{25 km}\end{tabular} &                                                                                        \\ \hline
4                                                                                                   &              11.0                                                 &                                    26.6               &             73.2                                       &                         \textbf{99.1}                         &       2.13                                                                            \\
8                                                                                                   &      12.8                                                          &                                      31.3            &                74.9                                   &                    98.7                              &            1.80                                                                       \\
16                                                                                                  &     \textbf{13.5}                                                       &                          \textbf{33.1}                             &        \textbf{77.4}                                             &                     98.7                                 &              \textbf{1.66}               \\ \hline
\end{tabular}
}
\label{tab:tsl}
\end{table}

\section{Analysis of Predictions by Video Length}
\label{sec:vidlen}
The Mapillary(MSLS)\cite{warburg2020mapillary} dataset consists of sequences of lengths ranging from 16 frames to 494 frames. Our model is length agnostic, as it can accommodate video sequences of any length as per the design of our dataloader and the model architecture itself. During evaluation, we do not sample frames for any video sequence and feed them as they are. Thus our model can accommodate videos from any dataset, be it GaMa\cite{vyas2022gama} and CityGuessr68k\cite{kulkarni2024cityguessr} which have a fairly consistent length of approximately 40 and 100 frames respectively, or MSLS which has sequences of variable length. However, for a model to be robust to variable length, the performance of the model should not change with video length. 

\begin{figure}[t]
    \centering
    \includegraphics[width=0.48\textwidth]{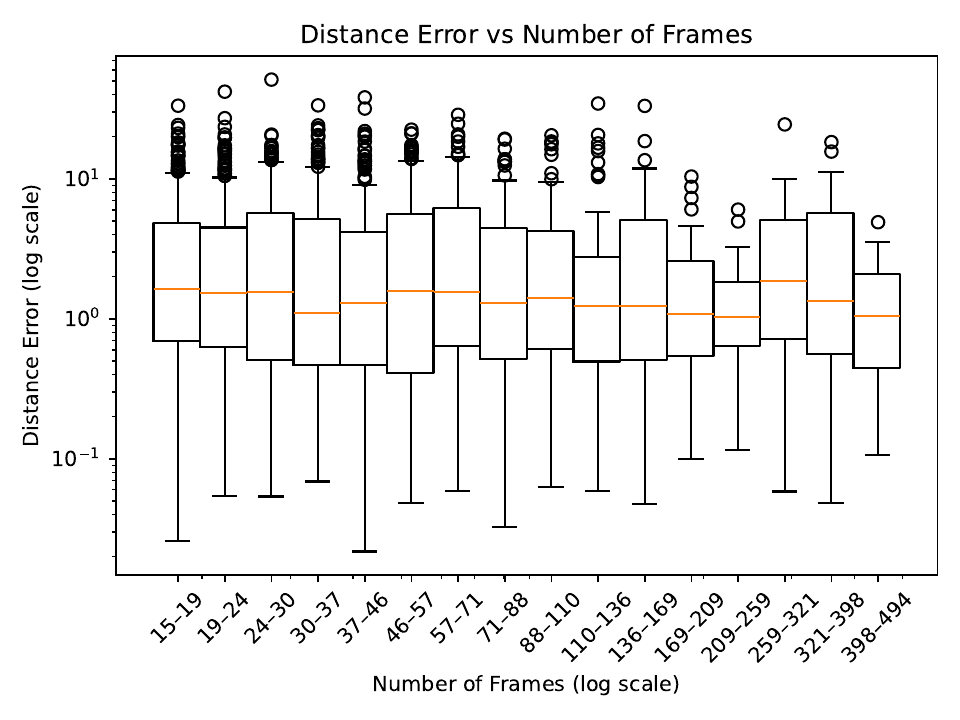}
    \caption{\textbf{Distance Error vs Number of Frames.} Median error is consistent, showing robustness of our model.}
    \label{fig:framesplot}
\end{figure}

To validate the robustness of our model, we compare the performance of our model across sequences of varying lengths from MSLS in terms of Video Distance Error, i.e., distance between centroid of the ground truth sequence and the prediction closest to the centroid of the predicted sequence(computation procedure discussed in Section \ref{sec:vidacc}). Fig. \ref{fig:framesplot} shows a boxplot of distance error vs frame length, with an orange line indicating the median error. We observe that the line does not deviate much implying the median distance error is largely independent of the no. of frames, and thus our model is robust to sequence length variation.

\begin{table*}[t]
\caption{Comparison of Uniform Grid Gallery evaluation and Validation-set Gallery evaluation on Mapillary(MSLS) dataset}
\resizebox{\linewidth}{!}{
\begin{tabular}{c|ccccc|ccccc|c|c}
\hline \hline
\multirow{4}{*}{\textbf{Model}} & \multicolumn{5}{c|}{\textbf{Uniform Grid Gallery Evaluation}}                                                                                                     & \multicolumn{5}{c|}{\textbf{Validation-set Gallery Evaluation}}                                                                                                         & \multirow{4}{*}{\textbf{$\Delta a_r$ {[}\%{]} @ 1 km}} & \multirow{4}{*}{\textbf{$\Delta$Median Distance}} \\ \cline{2-11}
                       & \multicolumn{4}{c|}{\textbf{Distance ($a_r$ {[}\%{]} @ km)}}              & \multirow{3}{*}{\begin{tabular}[c]{@{}c@{}}\textbf{Median}\\ \textbf{Distance}\\ \textbf{Error(km)}\end{tabular}} & \multicolumn{4}{c|}{\textbf{Distance ($a_r$ {[}\%{]} @ km)}}              & \multirow{3}{*}{\begin{tabular}[c]{@{}c@{}}\textbf{Median}\\ \textbf{Distance}\\ \textbf{Error(km)}\end{tabular}} &                                       &                                  \\
                       & \textbf{Sub-street}     & \textbf{Street} & \textbf{Locality} & \multicolumn{1}{c|}{\textbf{City}}  &                                                                                        & \textbf{Sub-street}     & \textbf{Street} & \textbf{Locality} & \multicolumn{1}{c|}{\textbf{City}}  &                                                                                        &                                       &                                  \\
                       & \textbf{0.5 km (500 m)} & \textbf{1 km}   & \textbf{5 km}     & \multicolumn{1}{c|}{\textbf{25 km}} &                                                                                        & \textbf{0.5 km (500 m)} & \textbf{1 km}   & \textbf{5 km}     & \multicolumn{1}{c|}{\textbf{25 km}} &                                                                                        &                                       &                                  \\ \hline
GeoCLIP-FineTuned      & 8.3            & 22.5   & 63.0     & \multicolumn{1}{c|}{93.9}  & 2.97                                                                                   & 19.9           & 31.1   & 66.3     & \multicolumn{1}{c|}{94.1}  & 2.30                                                                                   & 8.6                                   & 0.67                             \\
VidTAG (Ours)        & 21.5           & 41.0   & 76.7     & \multicolumn{1}{c|}{97.9}  & 1.35                                                                                   & 30.8           & 46.1   & 77.8     & \multicolumn{1}{c|}{96.3}  & 1.17                                                                                   & 5.1                                   & 0.18                             \\ \hline \hline
\end{tabular}
}
\label{tab:valgall}
\end{table*}

\section{Reference Gallery}
\label{sec:refgall}
Reference Gallery is an integral part of any retrieval based approach. As we propose a method for frame-to-GPS retrieval-based geolocalization, we have to construct an extensive gallery of GPS coordinates. GeoCLIP constructs its gallery by randomly sampling points from its training dataset (MP-16\cite{larson2017benchmarking}). This is not practical in our case, as the frame-wise video geolocalization problem requires higher amount of precision in location prediction as compared to its image counterpart, due to close proximity of frames of a video. An ideal gallery should enable a model to correctly identify its most appropriate match, while at the same time providing no information about the validation set. To achieve this delicate balance, we construct a uniform grid gallery based on the train set, which facilitates our model to perform fine-grained location prediction.  
\subsection{Uniform Grid Gallery Construction}
\label{sec:unigall}
We construct uniform grid galleries for both Mapillary(MSLS)\cite{warburg2020mapillary} and GaMa\cite{vyas2022gama} datasets. The procedure for gallery construction can be described as follows:
\begin{enumerate}
    \item For each region that contains a sizable amount of training data, get the maximum and minimum of both latitude and longitude($LAT_{MIN}$, $LON_{MIN}$, $LAT_{MAX}$, $LON_{MAX}$). (If there are a few outliers they can be skipped).
    \item To keep an error margin, determine a constant padding which is to be applied to ($LAT_{MIN}$, $LON_{MIN}$, $LAT_{MAX}$, $LON_{MAX}$). Also determine the resolution of the gallery (the finer the resolution the larger the gallery.
    \item Determine new 
    \\ $LAT_{MIN} = LAT_{MIN}-padding$, 
    \\ $LAT_{MAX}=LAT_{MAX}+padding$, 
    \\ $LON_{MIN}=LON_{MIN}-padding$, 
    \\ $LON_{MAX}=LON_{MAX}+padding$. 
    \item Compute distance($dis_{LAT}$) between ($LAT_{MIN}$, $LON_{MIN}$) and ($LAT_{MAX}$, $LON_{MIN}$), and distance($dis_{LON}$) between ($LAT_{MIN}$, $LON_{MIN}$) and ($LAT_{MIN}$, $LON_{MAX}$).
    \item Generate a grid with each point at a uniform distance (resolution), dividing the area into equidistant points in both directions
    \\ $N_{points} = (dis_{LAT}*dis_{LON})//resolution^2$
\end{enumerate}

\subsection{Evaluation with Val-set Gallery Retrieval}
\label{sec:valset}
A real world scenario entails not knowing the actual ground truth values pertaining to the queries, which is the reason for constructing a uniform grid gallery. However, if the model was to know the set of possible points already, the retrieval processs would be easier. To test this notion, instead of a blind retrieval from a grid gallery, we construct a gallery out of known set of GPS coordinates, which are the ground truth GPS coordinates corresponding to all frames in the validation set. Table \ref{tab:valgall} shows results of this experiment. We observe a rise in performance as compared to a grid gallery(blind) retrieval as hypothesized. However, this observed difference is significantly greater in GeoCLIP-Finetuned baseline, implying that our model does not rely on knowledge of the validation set. This demonstrates the robustness and generalization ability of our model.

\section{Video and Trajectory Metrics}
\label{sec:traj}
We use two metrics to measure the quality of a trajectory for performance evaluation of different models. Both metrics serve different purposes as they focus on different properties of a sequence to determine its quality and overall structure. In this section we highlight these metrics and discuss them below. We also outline the steps required to compute video level accuracy at distance thresholds.
\subsection{Video Distance Threshold Accuracy}
\label{sec:vidacc}
Frame-wise distance threshold accuracies are not sufficient for judging the performance of a model at video level. Thus we compute these for every video considering it as a single unit. This computation can be carried out as follows:
\begin{enumerate}
    \item Collect all model predictions and ground truth labels for a particular sequence
    \item Compute the centroid of the ground truth labels. This will serve as the ground truth label of the entire video sequence
    \item Compute the centroid of predictions pertaining to all frames in the sequence. Obtain the prediction that is closest to this centroid (this reduces error due to outliers). This will serve as the prediction for the entire video sequence
    \item Finally compute the distance between the prediction and the ground truth label. Compute distance accuracy at all thresholds using this distance measure.
\end{enumerate}

This enables us to effectively measure performance of a model at video level. However, this set of metrics cannot be used to judge the quality of formed trajectories. For this exact purpose we use DFD and MRD, which are described in the following subsections.

\subsection{Discrete Frechet Distance (DFD)}
\label{sec:dfd}
Discrete Frechet Distance(DFD)\cite{eiter1994computing}, also called Coupling distance is a variant of Frechet Distance which is defined as the length of the shortest straight line connecting two curves, that is sufficient for two individual points to traverse their respective curves from start to finish. If the curve is made of discrete points instead of a continuous function, the metric becomes Discrete Frechet Distance. Mathematically, given two curves A and B in the same metric space S, the Frechet Distance between them is given by,
\begin{align}
    & F(A, B) = \inf_{\alpha, \beta} \max_{t \in [0,1]} \{d(A(\alpha(t)), B(\beta(t)) \}
\end{align}
where d is the distance function of S, t is time reparametrized between [0,1] and $\alpha$ and $\beta$ are two continuous, non-decreasing surjections. 

Discrete Frechet Distance takes into account the flow of the two sequences, which makes it ideal to measure the quality of a predicted sequence of GPS coordinates against a true GPS trajectory.
\subsection{Mean Range Difference (MRD)}
\label{sec:mrd}
Mean Range Difference(MRD)\cite{gaharwar2023xi} is defined as the difference between the expanses covered by two curves, i.e., the range of values covered by the respective curves. Mathematically, given two curves A and B in the same metric space S, the Range Difference between them is given by,
\begin{align}
    & R(A, B) = abs(\{\max_{A}-\min_A\} - \{\max_{B}-\min_B\})  
\end{align}
where, abs is the absolute value function and max and min are maximum and minimum functions respectively. Mean Range Difference is the Mean of Range Differences over multiple sequences. It takes into account the overall scope of a sequence. MRD is primarily a 1D metric. However it can be adapted to 2D by computing MRD in both dimensions seperately and then taking their mean. This definition makes it similar to GIoU\cite{rezatofighi2019generalized} between two objects, but for curves. This makes MRD a good metric for measuring sequence quality. 

\begin{figure*}[!t]
  \centering
  \begin{subfigure}{0.37\linewidth}
    \centering
    \includegraphics[width = \textwidth]{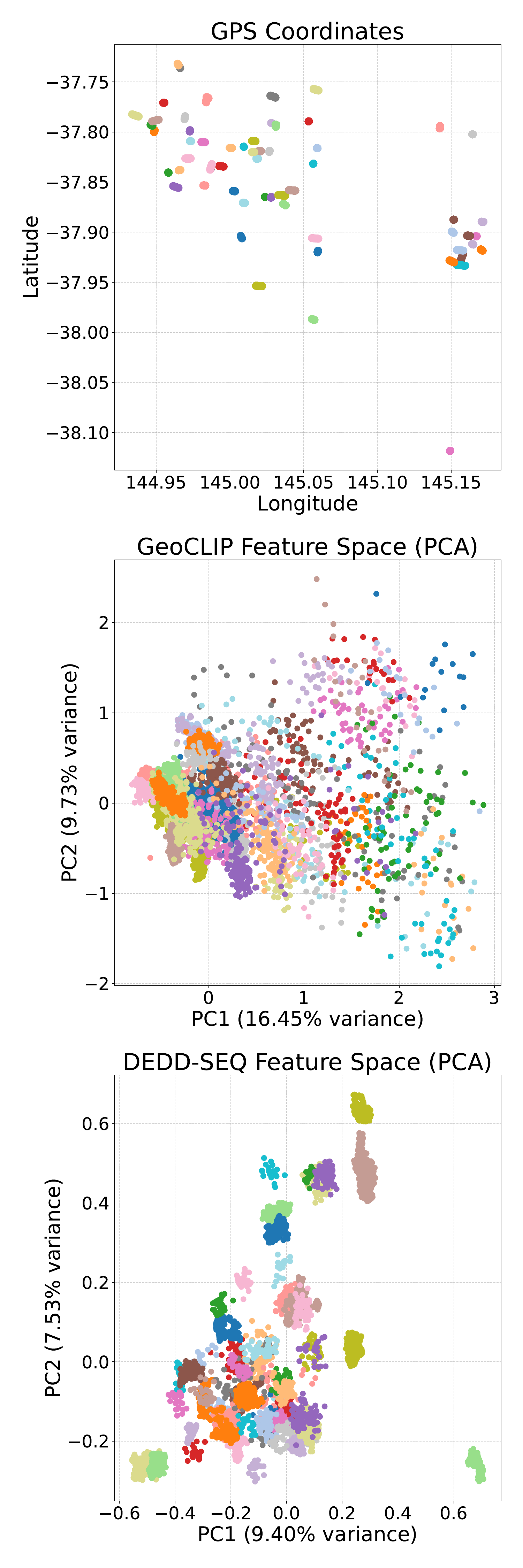}
    \caption{\textbf{Melbourne, Australia}}
    \label{fig:melbourne}
  \end{subfigure}
  \begin{subfigure}{0.37\linewidth}
    \centering
    \includegraphics[width = \textwidth]{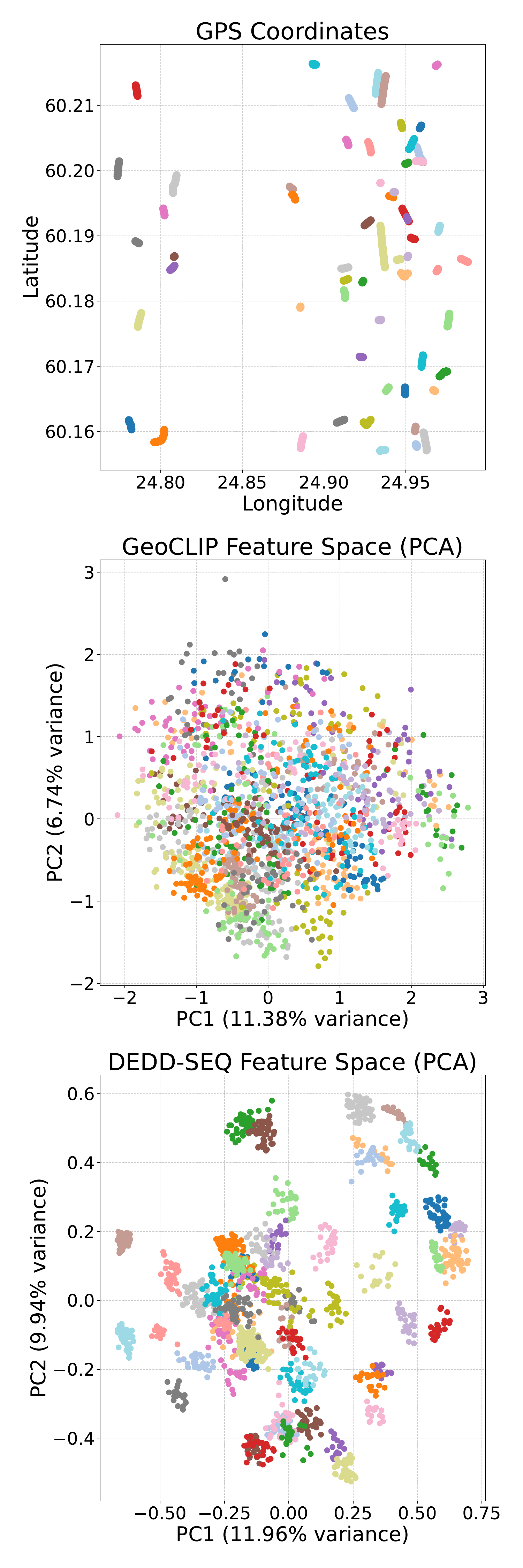}
    \caption{\textbf{Helsinki, Finland}}
    \label{fig:helsinki}
  \end{subfigure}
   \caption{\textbf{Feature Space Visualization comparison for output of GeoCLIP\cite{vivanco2024geoclip} and VidTAG.} Figure shows the feature analysis in the form of PCA plots for GeoCLIP and our model. First row shows The GPS spread of sequences in Melbourne and Helsinki respectively. Second row displays corresponding PCA plots for sequence-wise outputs of GeoCLIP, while row 3 displays the corresponding PCA plots for our model VidTAG. We see that the features of VidTAG are a lot more distinct as compared to GeoCLIP, whose features are more scattered and unorganized.}
   \label{fig:featviz}
\end{figure*}

\begin{table*}[t]
\caption{Performance of Unified Model on Mapillary(MSLS) and GAMa datasets}
\resizebox{\linewidth}{!}{
\begin{tabular}{l|ccccccc|ccccccc}
\hline  \hline
\multicolumn{1}{c|}{\multirow{4}{*}{\textbf{Model}}} & \multicolumn{7}{c|}{\textbf{Mapillary(MSLS)}}                                                                                                                                                                                                                                                                           & \multicolumn{7}{c}{\textbf{GAMa}}                                                                                                                                                                                                                                                                                       \\ \cline{2-15} 
\multicolumn{1}{c|}{}                                & \multicolumn{4}{c|}{\textbf{Distance ($a_r$ {[}\%{]} @ km)} $\uparrow$}                                        & \multicolumn{1}{c|}{\multirow{3}{*}{\textbf{\begin{tabular}[c]{@{}c@{}}Median\\ Distance\\ Error(km) $\downarrow$\end{tabular}}}} & \multicolumn{1}{c|}{\multirow{3}{*}{\textbf{DFD} $\downarrow$}} & \multirow{3}{*}{\textbf{MRD} $\downarrow$} & \multicolumn{4}{c|}{\textbf{Distance ($a_r$ {[}\%{]} @ km)} $\uparrow$}                                        & \multicolumn{1}{c|}{\multirow{3}{*}{\textbf{\begin{tabular}[c]{@{}c@{}}Median\\ Distance\\ Error(km) $\downarrow$\end{tabular}}}} & \multicolumn{1}{c|}{\multirow{3}{*}{\textbf{DFD} $\downarrow$}} & \multirow{3}{*}{\textbf{MRD} $\downarrow$} \\
\multicolumn{1}{c|}{}                                & \textbf{Sub-street}     & \textbf{Street}      & \textbf{Locality}    & \multicolumn{1}{c|}{\textbf{City}}  & \multicolumn{1}{c|}{}                                                                                                & \multicolumn{1}{c|}{}                              &                               & \textbf{Sub-street}     & \textbf{Street}      & \textbf{Locality}    & \multicolumn{1}{c|}{\textbf{City}}  & \multicolumn{1}{c|}{}                                                                                                & \multicolumn{1}{c|}{}                              &                               \\
\multicolumn{1}{c|}{}                                & \textbf{0.5 km (500 m)} & \textbf{1 km}        & \textbf{5 km}        & \multicolumn{1}{c|}{\textbf{25 km}} & \multicolumn{1}{c|}{}                                                                                                & \multicolumn{1}{c|}{}                              &                               & \textbf{0.5 km (500 m)} & \textbf{1 km}        & \textbf{5 km}        & \multicolumn{1}{c|}{\textbf{25 km}} & \multicolumn{1}{c|}{}                                                                                                & \multicolumn{1}{c|}{}                              &                               \\ \hline
\textit{\textbf{Dataset-specific Models}}                     &                         &                      & \multicolumn{1}{l}{} & \multicolumn{1}{l|}{}               & \multicolumn{1}{c|}{}                                                                                                & \multicolumn{1}{l|}{}                              & \multicolumn{1}{l|}{}         &                         &                      & \multicolumn{1}{l}{} & \multicolumn{1}{l|}{}               & \multicolumn{1}{c|}{}                                                                                                & \multicolumn{1}{c|}{}                              &                               \\
GeoCLIP-FineTuned                                    & 8.3                     & 22.5                 & 63.0                 & \multicolumn{1}{c|}{93.9}           & \multicolumn{1}{c|}{2.97}                                                                                            & \multicolumn{1}{c|}{22.52}                         & 2.82                          & 16.3                    & 28.3                 & 57.8                 & \multicolumn{1}{c|}{89.1}           & \multicolumn{1}{c|}{3.28}                                                                                            & \multicolumn{1}{c|}{6.50}                          & 0.50                          \\
VidTAG                                             & 21.5                    & 41.0                 & 76.7                 & \multicolumn{1}{c|}{97.9}           & \multicolumn{1}{c|}{1.35}                                                                                            & \multicolumn{1}{c|}{3.87}                          & 1.07                          & 35.4                    & 53.1                 & 77.8                 & \multicolumn{1}{c|}{94.4}           & \multicolumn{1}{c|}{0.88}                                                                                            & \multicolumn{1}{c|}{0.39}                          & 0.17                          \\ \hline
\textit{\textbf{Universal Models}}                            & \multicolumn{1}{l}{}    & \multicolumn{1}{l}{} & \multicolumn{1}{l}{} & \multicolumn{1}{l|}{}               & \multicolumn{1}{l|}{}                                                                                                & \multicolumn{1}{l|}{}                              & \multicolumn{1}{l|}{}         & \multicolumn{1}{l}{}    & \multicolumn{1}{l}{} & \multicolumn{1}{l}{} & \multicolumn{1}{l|}{}               & \multicolumn{1}{l|}{}                                                                                                & \multicolumn{1}{l|}{}                              & \multicolumn{1}{l}{}          \\
GeoCLIP-ZeroShot                                     & 0.9                     & 2.7                  & 22.9                 & \multicolumn{1}{c|}{84.6}           & \multicolumn{1}{c|}{11.54}                                                                                           & \multicolumn{1}{c|}{24.94}                         & 2.83                          & 4.1                     & 10.1                 & 33.5                 & \multicolumn{1}{c|}{74.9}           & \multicolumn{1}{c|}{11.15}                                                                                           & \multicolumn{1}{c|}{14.29}                         & 1.51                          \\
VidTAG (Unified)                                   & 15.3                    & 35.4                 & 75.3                 & \multicolumn{1}{c|}{98.8}           & \multicolumn{1}{c|}{1.63}                                                                                            & \multicolumn{1}{c|}{1.47}                          & 0.57                          & 19.9                    & 35.7                 & 66.5                 & \multicolumn{1}{c|}{91.9}           & \multicolumn{1}{c|}{1.95}                                                                                            & \multicolumn{1}{c|}{1.51}                          & 0.35                          \\ \hline \hline
\end{tabular}
}
\label{tab:unified}
\end{table*}

\section{Training of Baselines/SoTA models}
\label{sec:sota}
We compare our results to 4 types of baselines. We finetune all models for fair comparison (For GeoCLIP, we show both ZeroShot and FineTuned performance as it is our primary baseline). Types of baselines and their training pipelines are described in the following subsections.
\subsection{Zero Shot}
\label{sec: zs}
We show Zero Shot performance on GeoCLIP to establish a level ground. As the model is evaluated Zero-Shot there is no training involved. However, we also train a version of GeoCLIP, which will be described in Section \ref{sec:gcft}.

\subsection{Classification-based SoTA models}
\label{sec:sotaclass}
We compare our model to 3 classification-based models, namely, PlaNet\cite{weyand2016planet}, ISNs\cite{muller2018geolocation} and GeoDecoder\cite{clark2023we}. For training such models, we require the data points to be divided into S2-cells. We take all points in the train split of Mapillary(MSLS)\cite{warburg2020mapillary} and using Google S2 Geometry library\footnote{\url{code.google.com/archive/p/s2-geometry-library}} divide all these points into S2 cells. For the finest granularity, we choose Level 13 S2 cells. PlaNet and ISNs are trained on these S2 classes. As GeoDecoder is a hierarchical model, we divide the training points into Level 8, 10 and 11 S2 cells in addition to the already computed Level 13. This resulted in 1400 cells at level 13, 337 cells at level 11, 146 cells at level 10 and 38 cells at level 8. For GaMa, we follow a similar procedure with levels 13, 10, 8 and 7, with 1882, 293, 94 and 56 S2 cells respectively. The respective models are trained according to their prescribed settings. We also compute scene labels for training ISNs and GeoDecoder using Places2\cite{zhou2017places}. The validation is performed by predicting an S2 cell, and computing distance from its cell center, as per the process for these methods. 
\subsection{Classification-based Foundation Baselines}
\label{sec:foundsota}
Along with the pre-established classification-based models, we also train more baselines. As we use CLIP and DINOv2 backbones in our model, we also compare with methods with these foundation models as a backbone. We finetune these models using procedure similar to other classification-based models with S2 cell classes as described in the previous section.

\subsection{Retrieval-based SoTA model}
\label{sec:gcft}
Finally, we also finetune GeoCLIP treating all frames as individual images for the network. For a fair comparison, we keep all settings exactly same as our model, and validate the model using the uniform grid gallery.

\section{Visualizations in Feature Space}
\label{sec:vizfeat}
Analyzing the features output by a model can provide further insight into the model's learning capability. The more distinct the features, the better the model's ability to recognize patterns and perform retrieval. To gauge our model in a similar light, we visualize the features of the predictions of our model. To generate these visualizations, geospatial data and feature embeddings are processed systematically. Precomputed feature vectors and corresponding GPS coordinates for a city, are loaded and concatenated into unified datasets for each model, while sequence boundaries are maintained using labeled indices. Dimensionality reduction is performed on the feature sets using Principal Component Analysis (PCA)\cite{pearson1901liii}, reducing their dimensions to two while preserving the maximum variance. Fig. \ref{fig:featviz} shows this visualization for sequences in Melbourne and Helsinki, where we see that our model is able to differentiate the features of each sequence, as seen in the bottom row. We compare this visualization with a similar PCA-reduced feature space of GeoCLIP\cite{vivanco2024geoclip}(middle row), and observe that GeoCLIP's feature space is much more scattered and less coherent. This demonstrates the effectiveness of the proposed components in aligning the features of the frames of a video with each other. 

To ensure clarity and interpretability, a subset of sequences is randomly selected for down-sampling, reducing the density of points in the plots while preserving representativeness. Color coding is applied consistently across the visualizations to indicate sequence membership, allowing for a direct comparison of geospatial coherence between GPS coordinates(top row) and the feature embeddings. These visualizations reveal a strong structural relationship between the geographic data and the latent feature 
spaces of our model.

\begin{table}[t]
\caption{Performance of Unified Model on CityGuessr68k}
\resizebox{0.48\textwidth}{!}{
\begin{tabular}{lcccc}
\toprule
\multicolumn{1}{c}{\textbf{Model}} & \multicolumn{1}{c}{\textbf{City $\uparrow$}} & \multicolumn{1}{c}{\textbf{State $\uparrow$}} & \multicolumn{1}{c}{\textbf{Country $\uparrow$}} & \multicolumn{1}{c}{\textbf{Continent $\uparrow$}} \\ \midrule

CityGuessr & 69.6 & 70.2 & 74.8 & 83.8                                   \\ 
VidTAG & 94.9 & 95.5 & 96.8 & 98.5                  \\
\midrule
VidTAG (Unified) & 79.0 & 80.2 & 89.1 & 95.0 \\
\bottomrule
\end{tabular}
}
\label{tab:unified-cg}
\end{table}

\section{Unified Model}
\label{sec:unified}

While VidTAG trained on GAMa and MSLS achieves commendable framewise (and video level on CityGuessr68k) geolocalization performance, the ultimate goal is to build a universal model capable of precise geolocalization without fine-tuning on a specific dataset. Given the limitations of the existing datasets we seek to achieve a limited version of this by training a single unified model to geolocalize all 3 video datasets at once.

We prepare a training split for the unified model by carefully combining training data from all 3 sources (Mapillary(MSLS), GAMa and CityGuessr68k) in the following way:
\begin{enumerate}
    \item As Mapillary(MSLS) is a smaller dataset, we take the entire training split.
    \item For CityGuessr68k, 
    \begin{enumerate}
        \item We remove training data from cities which are also present in MSLS to avoid false negatives in the contrastive loss.
        \item As CityGuessr68k is very large, we sample the data in such a way that the number of sequences is roughly equivalent to MSLS.
        \item This is achieved by taking all sequences from cities with less than 80 sequences, and randomly sampling 80 sequences each, from the rest.
    \end{enumerate}
    \item For GaMa, we randomly sample videos so that the corpus size is equivalent to the other two.
\end{enumerate}   

We train a model on this unified data for 200 epochs at an learning rate decay rate of 0.97. Rest of the settings are unchanged. 

Tables \ref{tab:unified} and \ref{tab:unified-cg} show evaluation performance of our model on individual datasets. Across all 3 datasets, we observe that our unified model outperforms state-of-the-art models by large margins. We see a 13\% improvement at 1km on MSLS, and 7\% improvement at 1km on GaMa. We also observe significantly lower median distance error, DFD and MRD in both cases. The unified model also outperforms the best model on CityGuessr68k  by 10\% at City level. However, there is still some gap to close between the unified model and VidTAG trained on individual datasets.

\section{Computational Cost}
\label{sec:compute}

We provide training time (on A6000s) and inference compute cost in Table~\ref{tab:comp}. VidTAG achieves significantly better performance at similar costs as baseline methods. In the Main paper Section \ref{sec:latency}, we also show that our model achieves a significant gain in performance as compared to GeoCLIP\cite{vivanco2024geoclip} with only a minimal decrease in throughput.

In the table, \textit{pre-computed backbone features} indicate computing and storing the CLIP\cite{radford2021learning} and DINOv2\cite{oquab2023dinov2} encoder features, and inputting them for finetuning the other components of the model. This does not affect the weights of the model as the encoder backbones are frozen anyway, while drastically reducing the training time.
\begin{table}[t]
\caption{Efficiency Analysis of proposed approach}
\centering
\small
\setlength{\tabcolsep}{2pt} 
\renewcommand{\arraystretch}{0.8}
\begin{tabular}{lcrr}
\toprule
\textbf{Model} & \textbf{Test GMACs} & \textbf{Train Time} & \textbf{Acc@1km} \\ \midrule
GeoCLIP-ZS           & \multirow{3}{*}{81.91} & 272 hrs & 2.7 \\
GeoCLIP (FT)  &  & 169 hrs & 22.5 \\
GeoCLIP (total)  &  & 441 hrs & 22.5 \\
Ours              & 83.47 & 68 hrs  & 41.0 \\
\midrule
\multicolumn{3}{l}{\textit{with pre-computed backbone features}} \\

GeoCLIP-ZS           & - & 2.5 hours & 2.7 \\
GeoCLIP (FT)  &  - & 2.5 hrs & 22.5 \\
GeoCLIP (total)  &  & 5 hrs & 22.5 \\
Ours              &  - & 2.75 hrs & 41.0 \\

\bottomrule
\end{tabular}
\label{tab:comp}
\end{table}

\begin{figure*}[h]
  \centering
  \begin{subfigure}{0.284\linewidth}
    \centering
    \includegraphics[width = \textwidth]{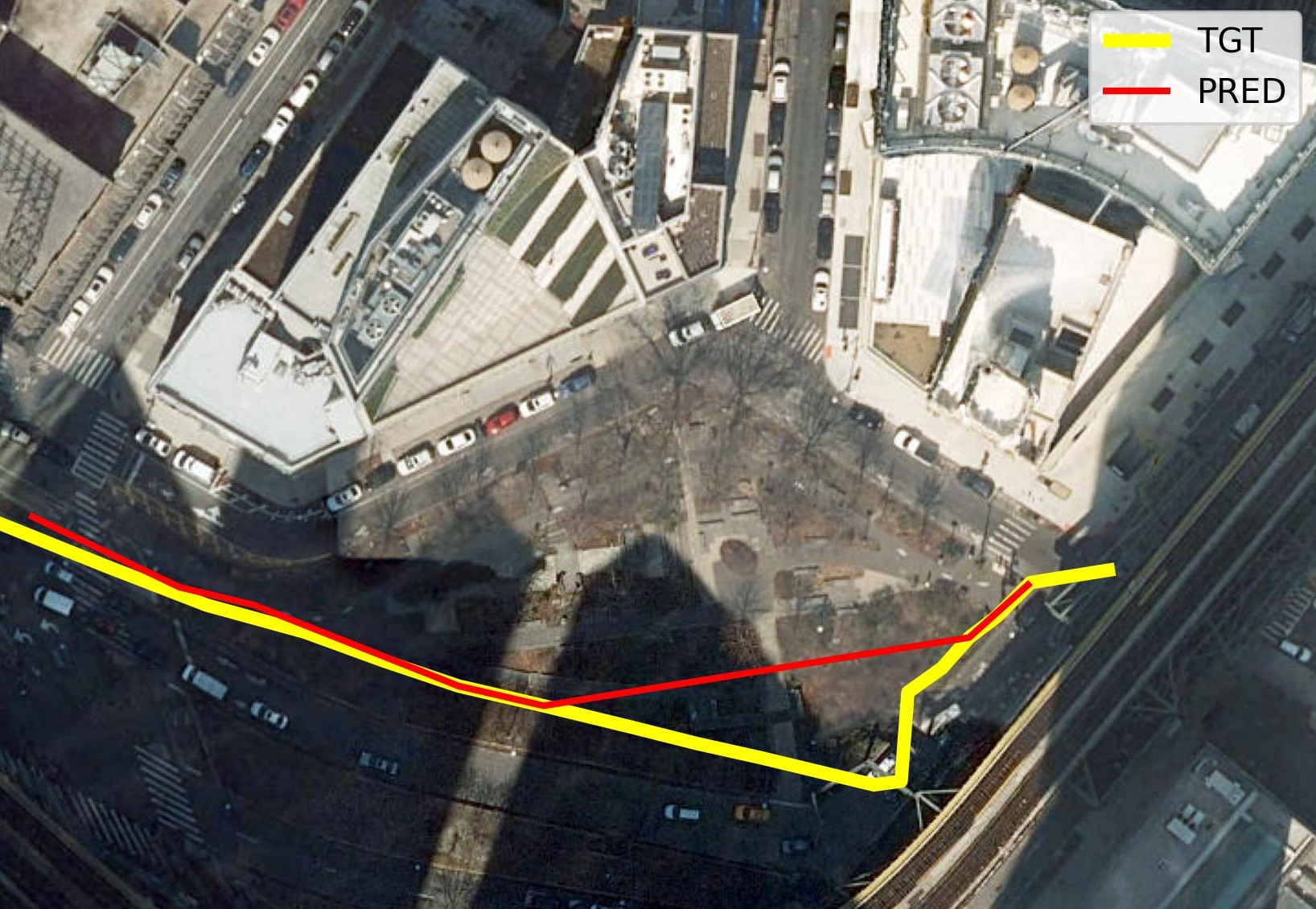}
    \caption{Brooklyn, NYC}
    \label{fig:perf1}
  \end{subfigure}
  \begin{subfigure}{0.183\linewidth}
    \centering
    \includegraphics[width = \textwidth]{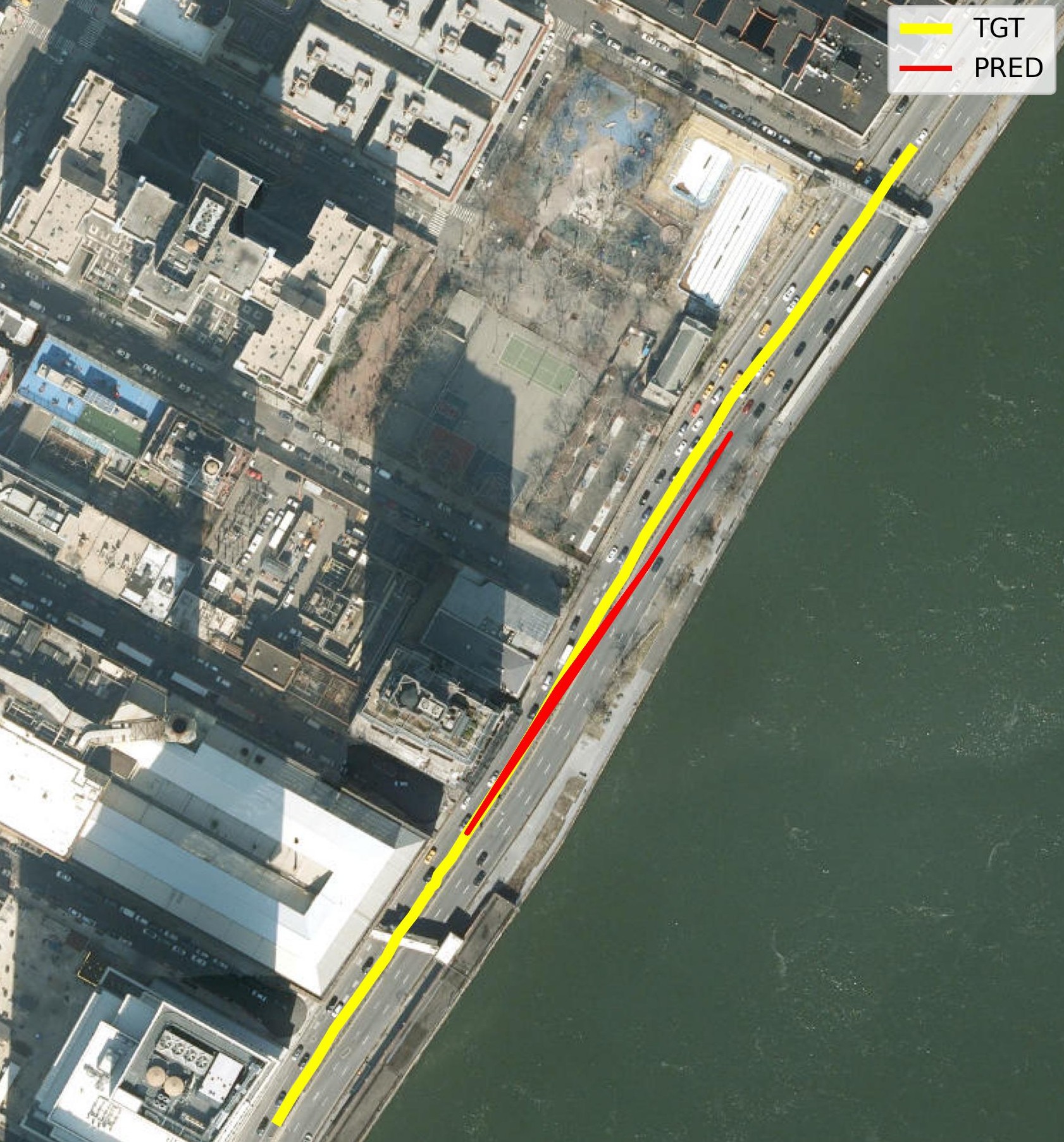}
    \caption{Manhattan, NYC}
    \label{fig:perf2}
  \end{subfigure}
  \begin{subfigure}{0.173\linewidth}
    \centering
    \includegraphics[width = \textwidth]{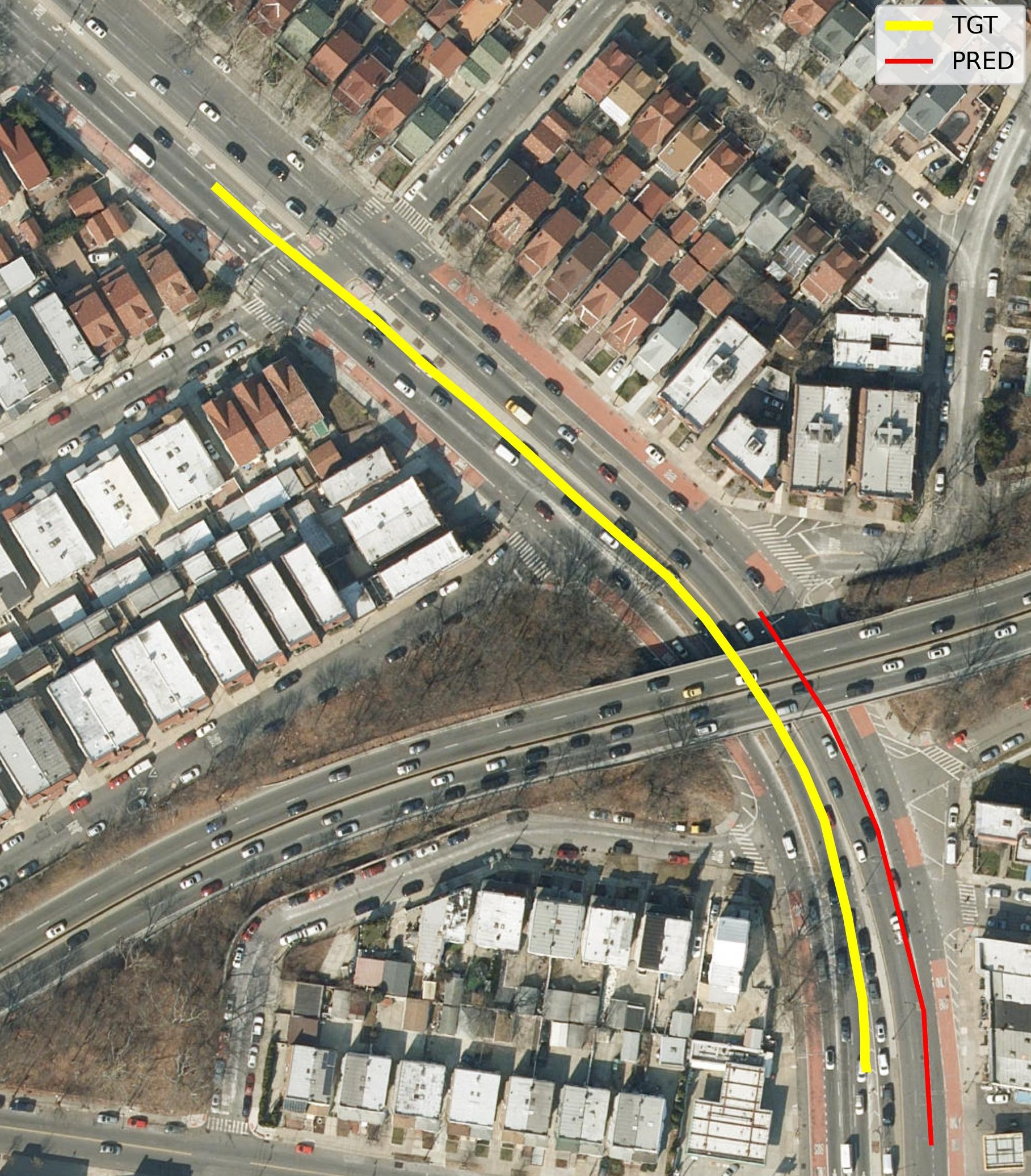}
    \caption{Queens, NYC}
    \label{fig:perf3}
  \end{subfigure}
  \begin{subfigure}{0.325\linewidth}
    \centering
    \includegraphics[width = \textwidth]{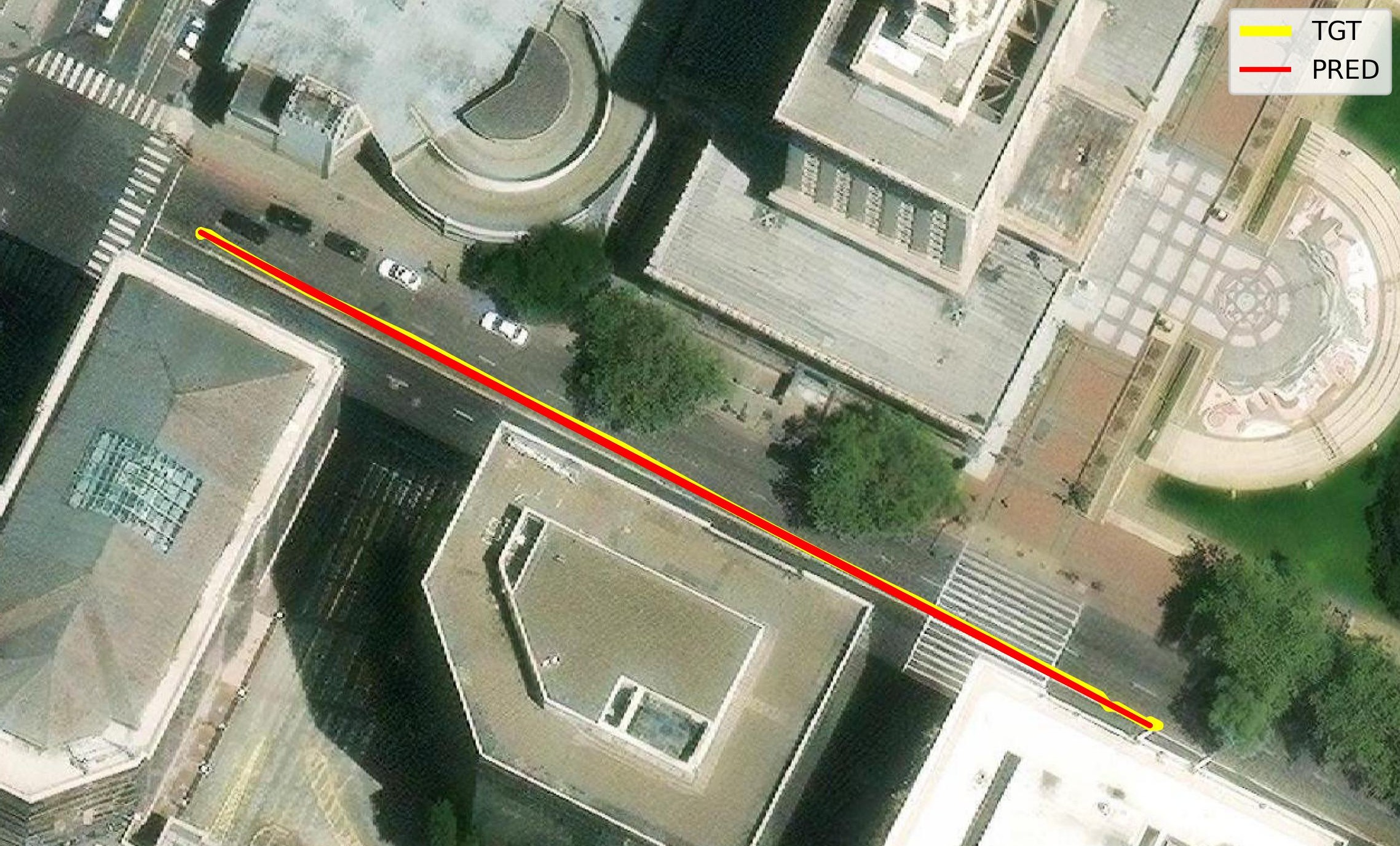}
    \caption{Oakland, SF}
    \label{fig:perf4}
  \end{subfigure}
  \begin{subfigure}{0.27\linewidth}
    \centering
    \includegraphics[width = \textwidth]{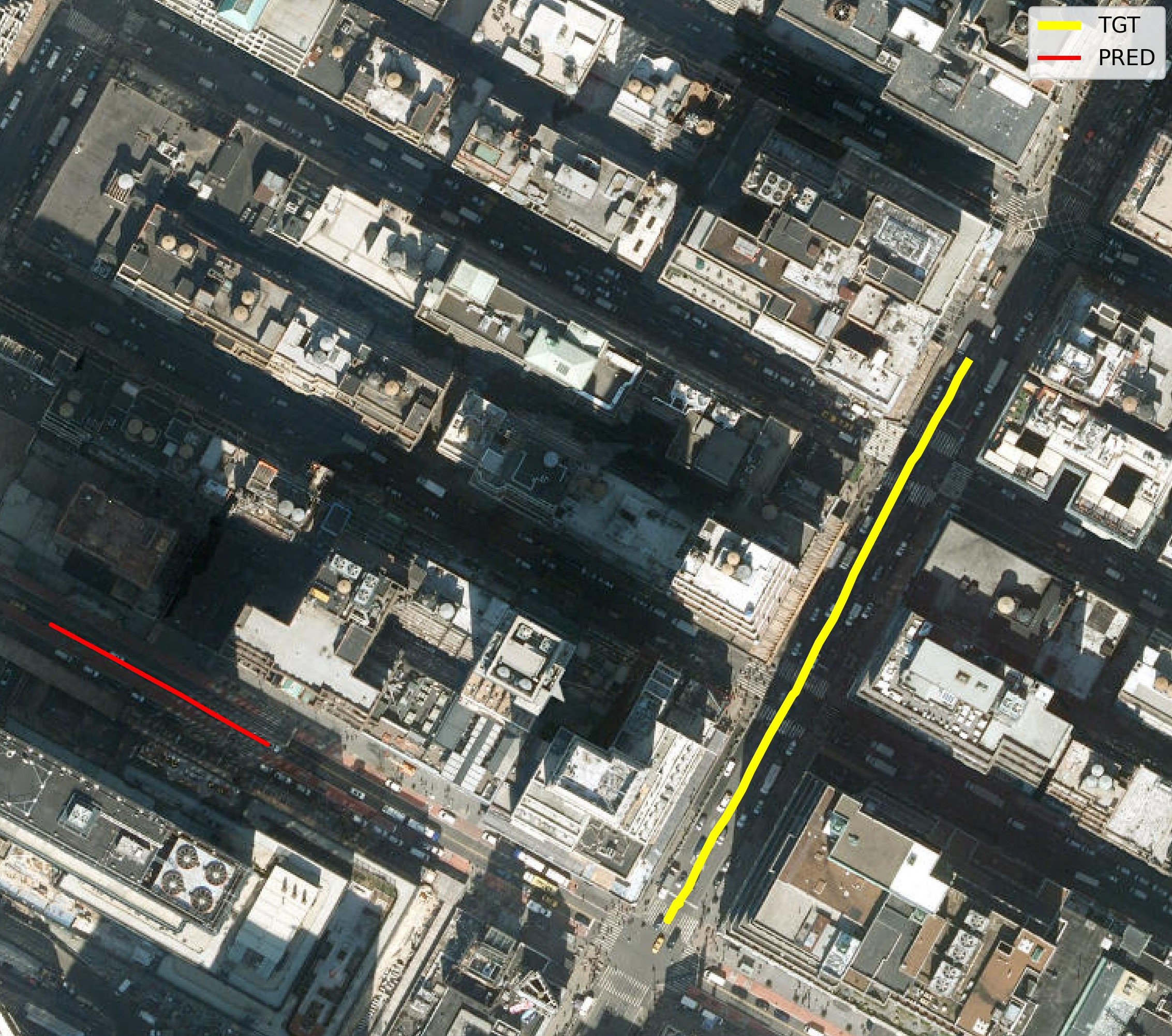}
    \caption{Manhattan, NYC}
    \label{fig:close1}
  \end{subfigure}
  \begin{subfigure}{0.24\linewidth}
    \centering
    \includegraphics[width = \textwidth]{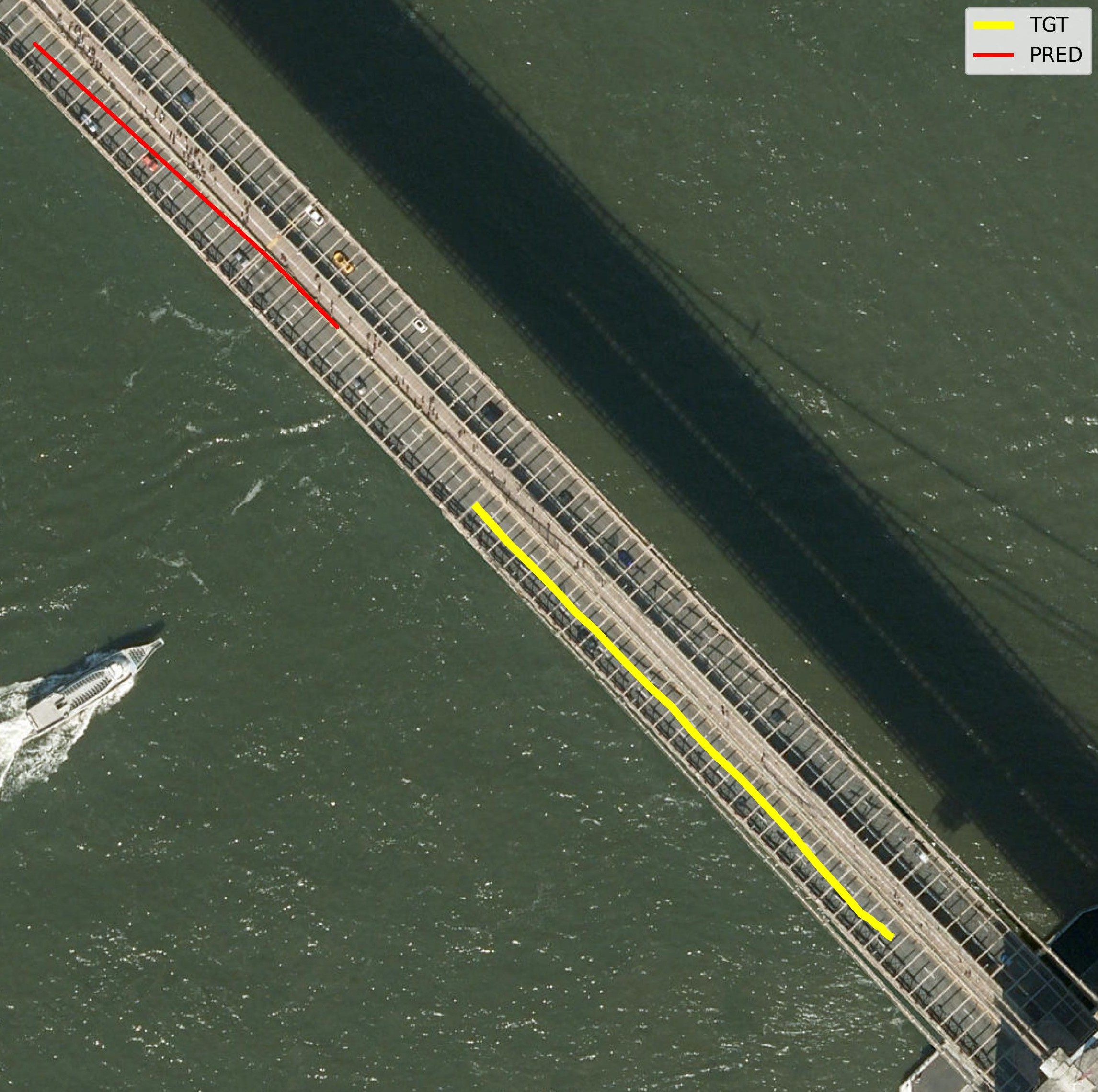}
    \caption{Brooklyn Bridge, NYC}
    \label{fig:close2}
  \end{subfigure}
  \begin{subfigure}{0.147\linewidth}
    \centering
    \includegraphics[width = \textwidth]{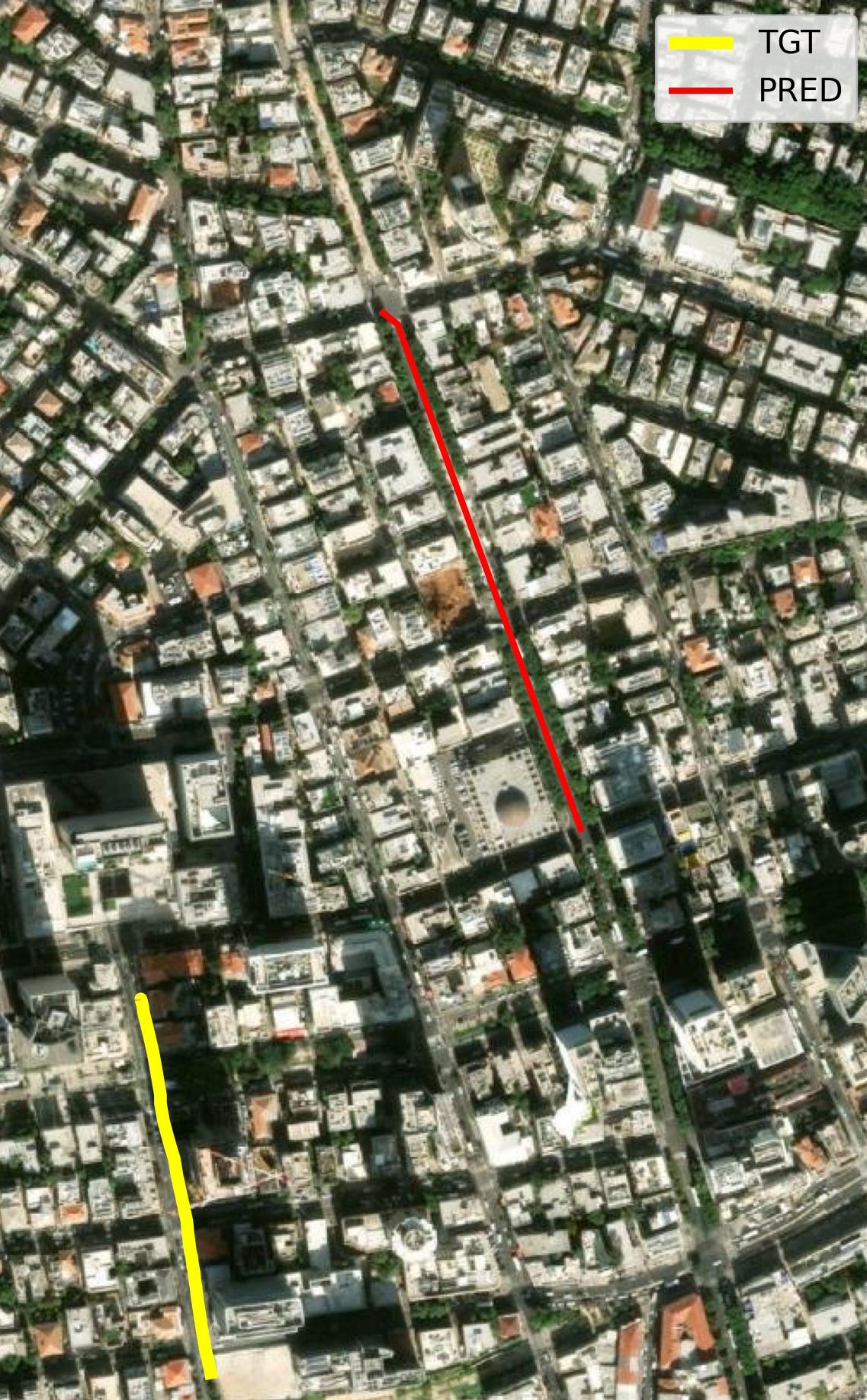}
    \caption{Tel Aviv, Israel}
    \label{fig:close3}
  \end{subfigure}
  \begin{subfigure}{0.31\linewidth}
    \centering
    \includegraphics[width = \textwidth]{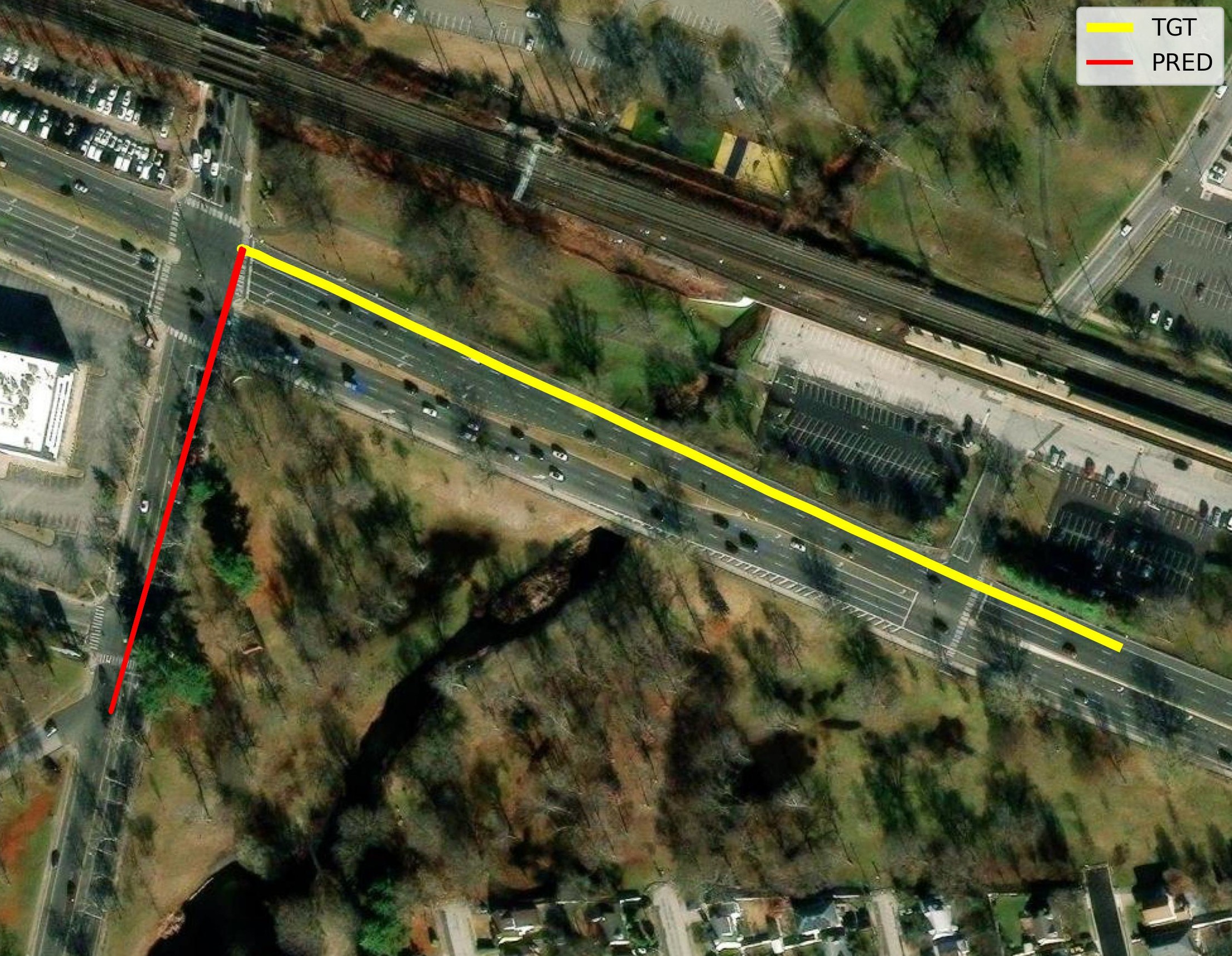}
    \caption{Long Island, NYC}
    \label{fig:close4}
  \end{subfigure}
   \caption{Qualitative Examples for trajectories of sequences in GaMa\cite{vyas2022gama} dataset. Row 1 shows almost perfectly predicted trajectories, while row 2 shows coherent trajectories predicted close to the actual trajectory.}
   \label{fig:qualgama}
\end{figure*}

\begin{figure*}[!h]
  \centering
  \begin{subfigure}{0.72\linewidth}
    \centering
    \includegraphics[width = 0.98\linewidth]{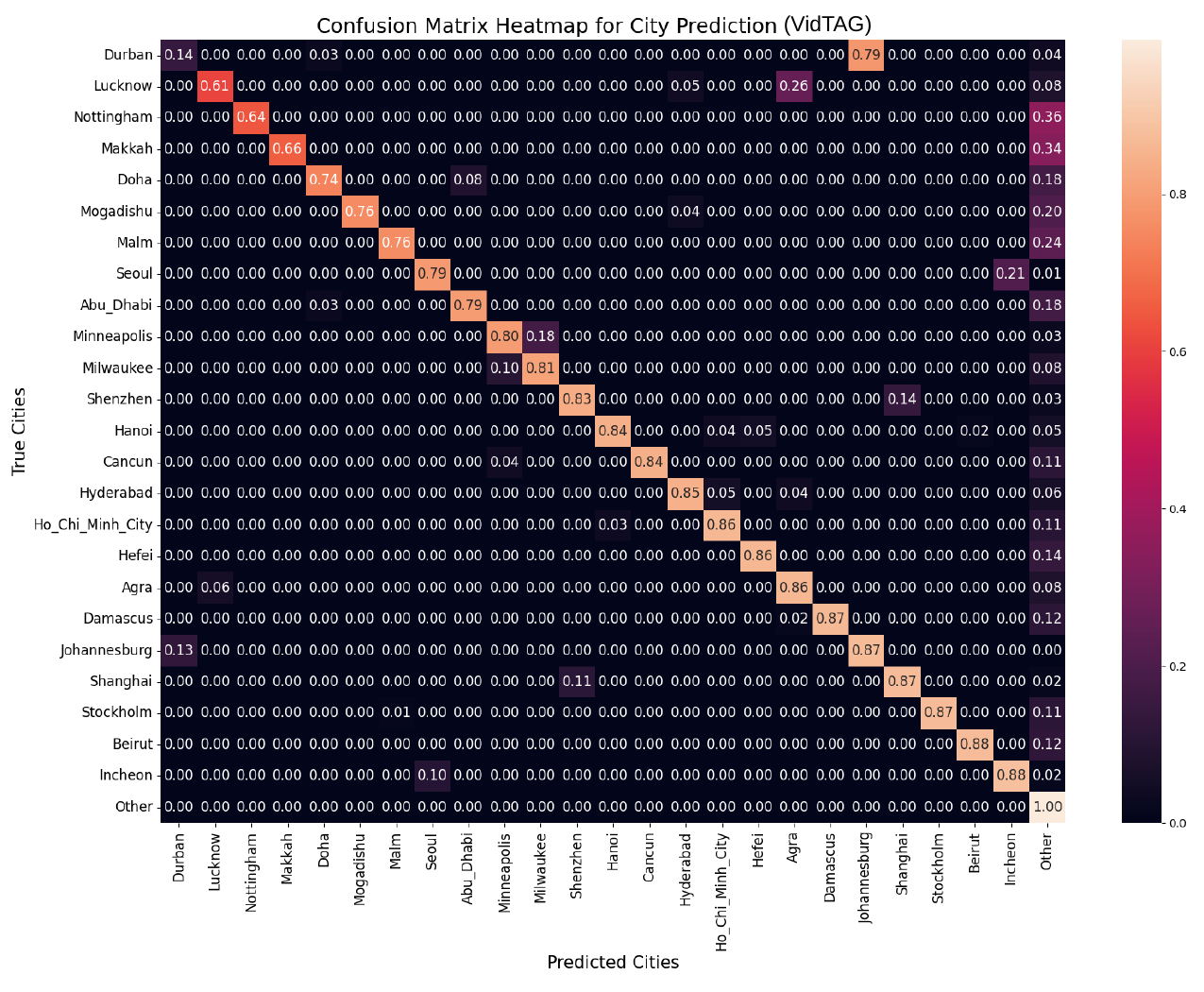}
    \label{fig:cm-deddseq}
  \end{subfigure}
  \begin{subfigure}{0.72\linewidth}
    \centering
    \includegraphics[width = 0.98\linewidth]{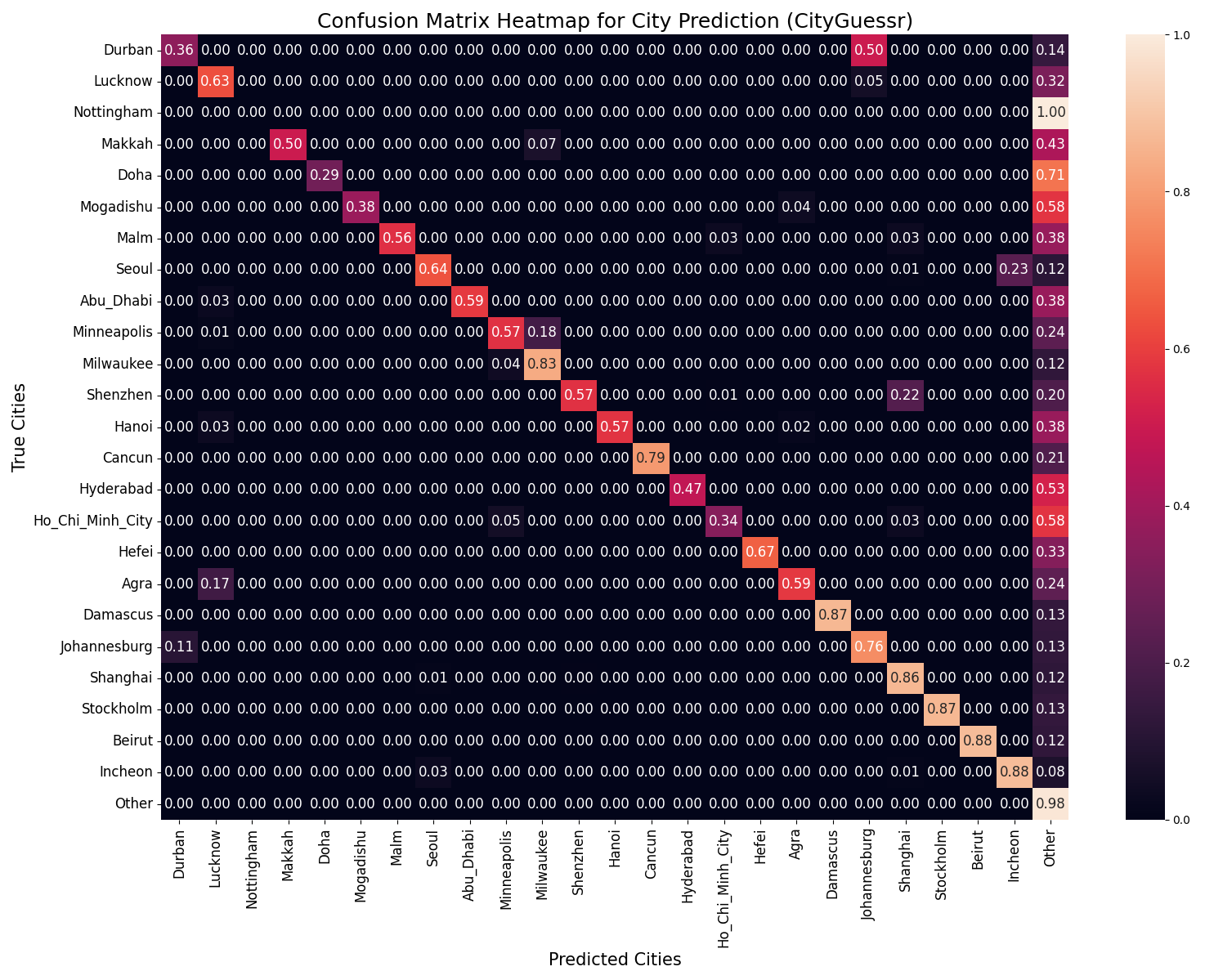}
    \label{fig:cm-cg}
  \end{subfigure}
   \caption{Figure shows confusion matrix heatmaps for 24 most difficult cities in CityGuesr68k for models VidTAG and CityGuessr respectively. It shows that our model can distinguish cities better, in cases where CityGuessr struggles. Also we see that both models mistake certain pairs of cities for each other, but CityGuessr in some cases confuses them with other cities, which does not happen in case of our model.}
   \label{fig:cm}
\end{figure*}

\begin{figure*}
    \centering
    \includegraphics[width=0.95\linewidth]{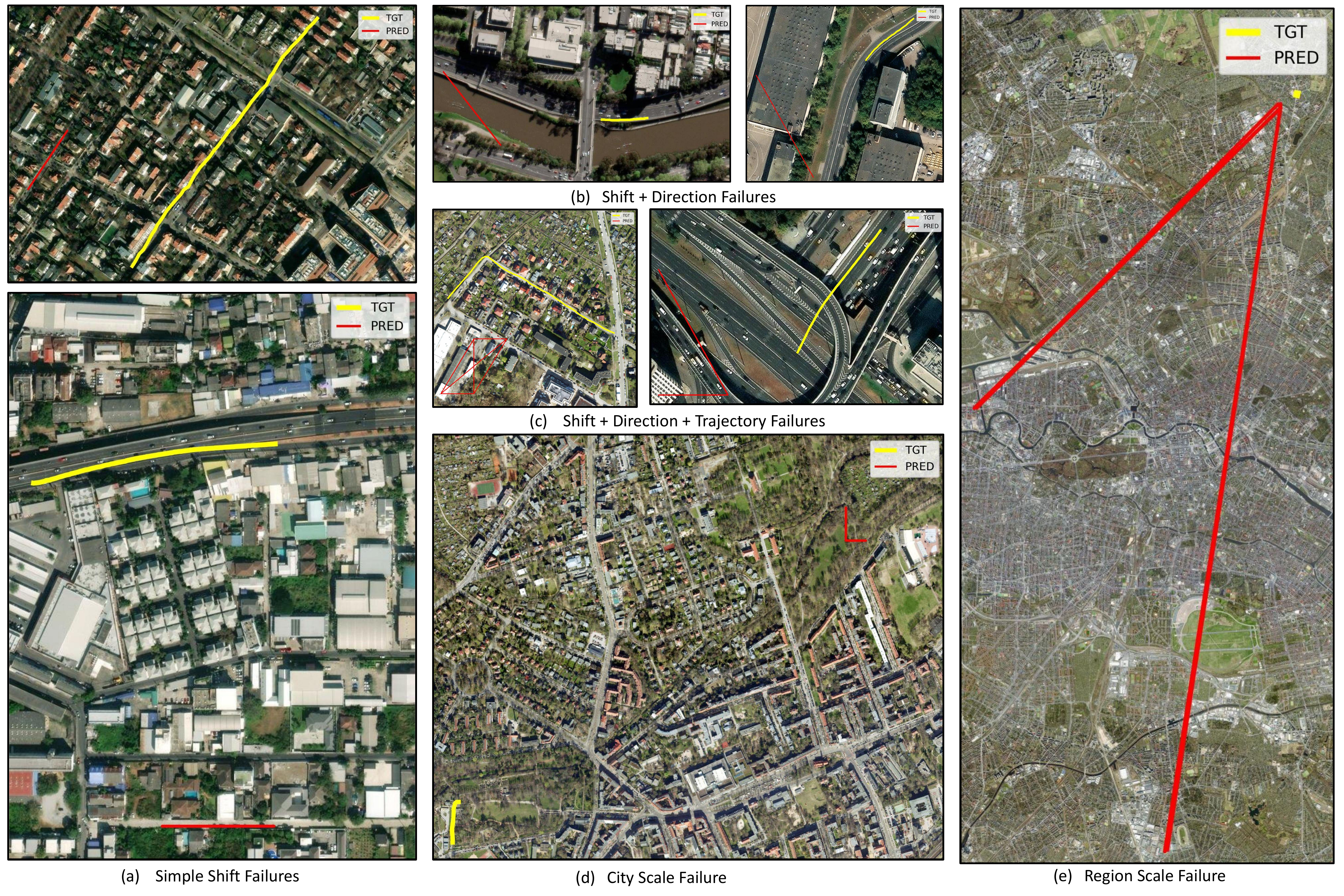}
    \caption{\textbf{Supplemental error analysis of \emph{VidTAG}.} (a) through (c) Breakdown of common model error types, highlighting dominant failure modes. (d) and (e) different scales of error: predictions are within the locality in vast majority of the cases, is in a different part of the city approximately 20\% of the time. Region level errors happen in $<$2\% of cases.}\label{fig:deddseq-error-analysis}
    \vspace{-1em}
\end{figure*}

\section{Qualitative Results on GaMa}
\label{sec:qualgama}
In Section \ref{sec:qual} of the main paper, we have shown a couple of qualitative examples of trajectories formed by the GPS coordinates predicted by our model for sequences in the Mapillary(MSLS)\cite{warburg2020mapillary} dataset. Here, we show a few sample trajectories for sequences in the GaMa\cite{vyas2022gama} split of BDD100k\cite{yu2020bdd100k} dataset, in Fig. \ref{fig:qualgama}. The first row shows examples of almost perfectly predicted trajectories from different locations. Row 2 shows a few examples where the model was unable to perfectly predict a trajectory, but predicted GPS coordinates were very close and form a coherent trajectory.

Fig. \ref{fig:perf1} and Fig. \ref{fig:perf3} shows correctly predicted trajectories that involve a curve/turn, while Fig. \ref{fig:perf2} and Fig. \ref{fig:perf4} show accurately predicted trajectories for sequences that are in a straight line.

Looking at the closely predicted trajectories(Row 2), we see that the model is able to recognize the structure of the trajectory, but fails to perfectly geolocalize points in the sequence. Fig. \ref{fig:close1} and Fig. \ref{fig:close4} show examples where the predicted trajectory is extremely close, but the model makes an error in the orientation,  while examples in Fig. \ref{fig:close2} and Fig. \ref{fig:close3} depict cases where model gets the orientation correct and is slightly off in terms of GPS location. We also see that the model is able to generate coherent trajectories across different locations and is not biased towards any specific place or city.

\section{City-level Geolocalization Performance Analysis on CityGuessr68k}
\label{sec:city}
As shown in Main paper Section \ref{sec:quant}, we outperform CityGuessr\cite{kulkarni2024cityguessr} as well as GeoCLIP\cite{vivanco2024geoclip} on the task of City Prediction on the CityGuessr68k dataset. As this task was originally designed as a classification task, we modified it to fit our retrieval framework by assigning the city center GPS coordinates as the ground truth annotations to all frames of videos from that city, as described in the main paper (Section \ref{sec:quant}). Here, we show more analysis regarding the same. Fig. \ref{fig:cm} shows a comparison between performance of VidTAG and CityGuessr in the form of confusion matrix heatmaps on the 24 most confusing cities, i.e., the cities on which the model gives the lowest performance, namely Durban, Lucknow, Nottingham, Makkah, Doha, Mogadishu, Malm, Seoul, Abu Dhabi, Minneapolis, Milwaukee, Shenzhen, Hanoi, Cancun, Hyderabad, Ho Chi Minh City, Hefei, Agra, Damascus, Johannesburg, Shanghai, Stockholm, Beirut and Incheon in decreasing order of difficulty.

We observe that overall, our model shows better ability to identify the city correctly, even in harder cases. We also notice some interesting patterns in case of a few pairs of cities which are mistaken for each other. A few examples are:
\begin{itemize}
    \item Durban and Johannesburg
    \item Lucknow and Agra
    \item Minneapolis and Milwaukee
    \item Shenzhen and Shanghai
    \item Incheon and Seoul
\end{itemize}

All these pairs of cities are located in the same country(in some cases within the same state/province/region) and have areas that look very similar, which might explain the models' confusion. However, we see that in most of these cases, VidTAG confuses cities in these pairs for each other, but quite a few times, CityGuessr confuses them for something completely different. All these observations demonstrate the capability of our model for the task of city prediction.

\section{Limitations and Failure Analysis}
\label{sec:limit}

Our model outperforms the best performing models on framewise video geolocalization across datasets of varying granularities. We also show our model's capability to address temporal inconsistency issue for the same. However, our model does have certain limitations. Firstly, the performance of our model is dependent on the resolution of the gallery. A coarser gallery leads to a subpar performance, while a finer gallery increases retrieval time. Secondly, even after achieving state-of-the-art performance on as fine threshold as 0.5km, a higher performance on an even finer threshold is desirable to achieve our goal of attaining perfect, temporally consistent trajectories. We aim to address these concerns in the future and reduce failures of our model. We provide some examples of failure cases. Fig.~\ref{fig:deddseq-error-analysis} shows that VidTAG’s errors are dominated by near-miss confusions among visually similar images from the same locale (Panels a–c). In terms of geographic scale (Panels d,e), mistakes remain local in the vast majority of cases; roughly 20\% fall in a different part of the same city, region-level errors are rare ($<\!2\%$). This pattern suggests the model is reasonably spatially calibrated and that in practical settings, lightweight, locality-aware post-processing could convert many near-misses into correct predictions while keeping catastrophic errors uncommon.
\vspace{-1em}
\section{Examples of Localizations}
\label{sec:loc}
\begin{figure*}[h]
    \centering
    \includegraphics[width=0.98\linewidth]{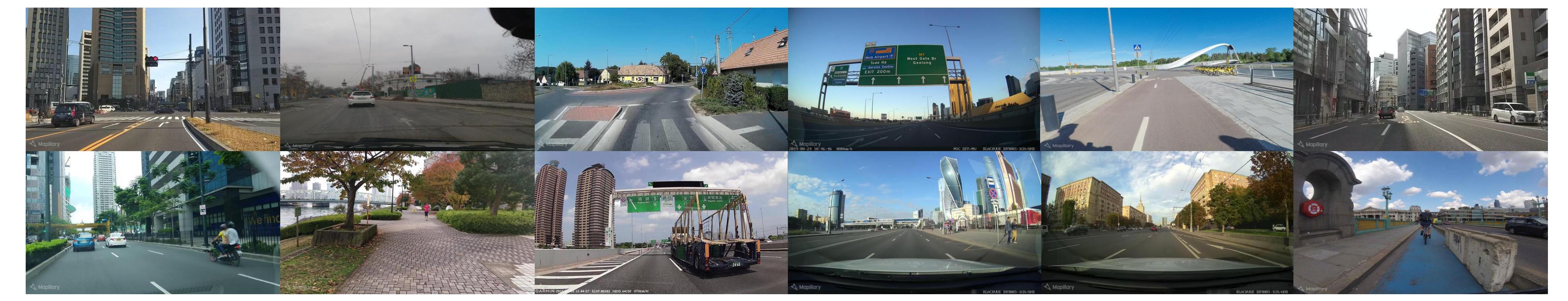}
    \caption{Visualization of samples for which our model correctly predicts location within 500 m (0.5 km). }
    \label{fig:substreet}
\end{figure*}

\begin{figure*}
    \centering
    \includegraphics[width=0.98\linewidth]{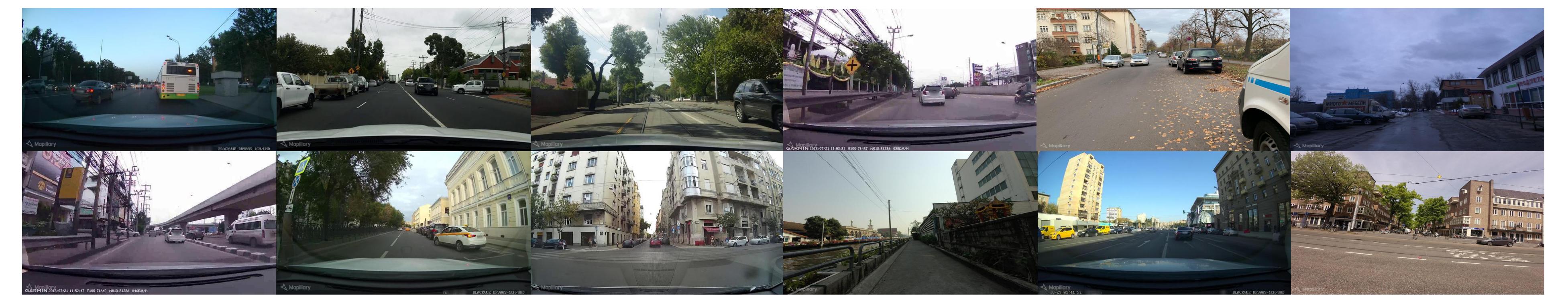}
    \caption{Visualization of samples for which our model correctly predicts location within 1 km but not within 500m (0.5 km). }
    \label{fig:street}
\end{figure*}

\begin{figure*}
    \centering
    \includegraphics[width=0.98\linewidth]{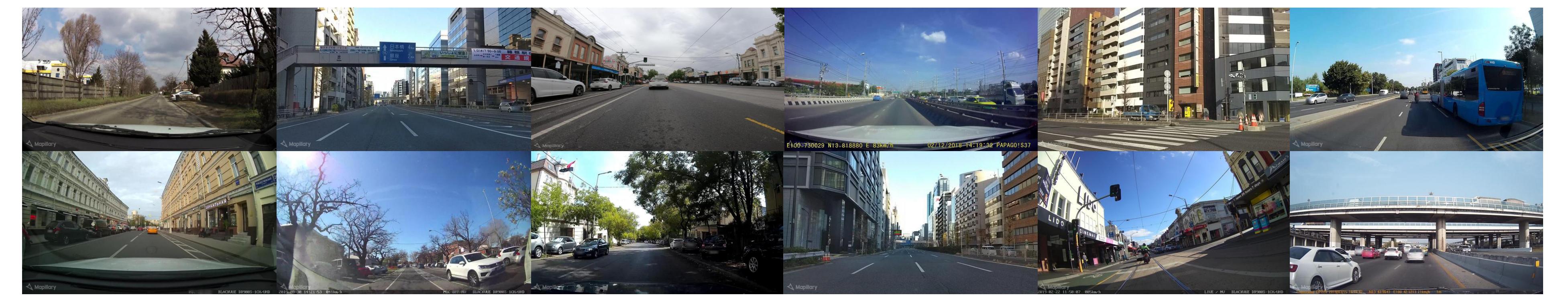}
    \caption{Visualization of samples for which our model correctly predicts location within 5 km but not within 1 km. }
    \label{fig:locality}
\end{figure*}

\begin{figure*}
    \centering
    \includegraphics[width=0.98\linewidth]{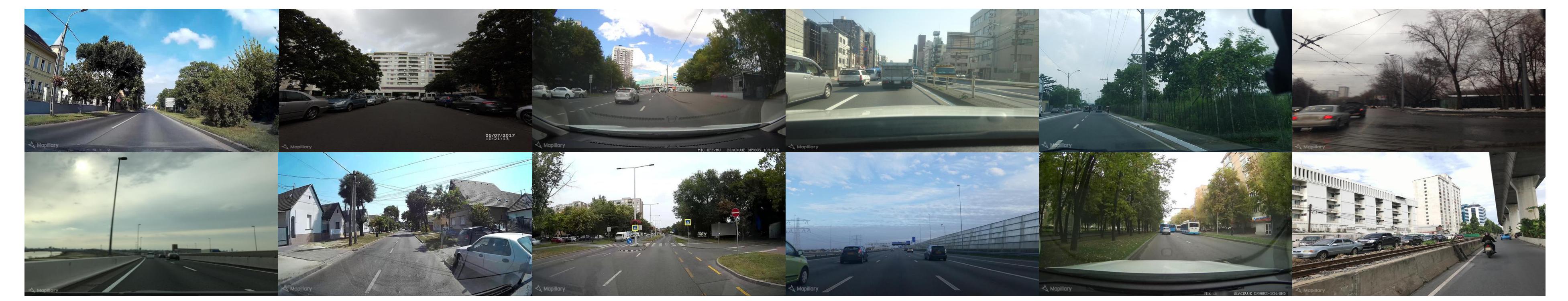}
    \caption{Visualization of samples for which our model correctly predicts location within 25km but not within 5km.}
    \label{fig:city}
\end{figure*}

\begin{figure*}
    \centering
    \includegraphics[width=0.98\linewidth]{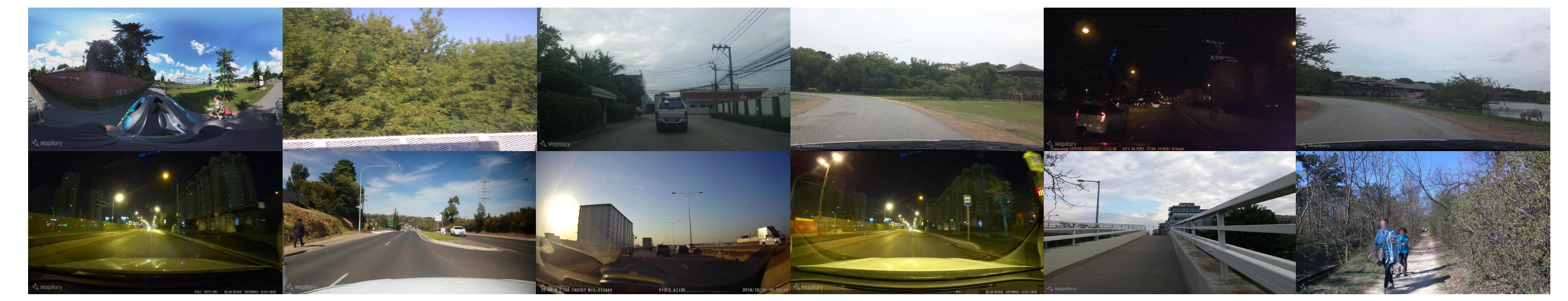}
    \caption{Visualization of samples for which our model cannot predict the location within 25 km. }
    \label{fig:fail}
\end{figure*}

As the performance of our model is evaluated at different distance thresholds, we provide a qualitative perspective, showcasing frames from multiple videos that have been accurately geolocalized within a certain threshold. We show 5 such cases. Fig. \ref{fig:substreet} shows frames from some videos that have been accurately geolocalized within 500 m (0.5 km) of the actual location, i.e. at the Sub-street level. Fig. \ref{fig:street} shows a few samples which have been geolocalized within 1 km by our model, i.e. at Street level, but not within 500 m. Fig. \ref{fig:locality} shows examples which have been successfully geolocalized at the Locality level, within 5 km, but cannot be recognized at the street level. Fig. \ref{fig:city} shows frames of videos that our model has localized within 25 km, i.e. in the same city as the actual location, but is not able to recognize its locality. Finally, Fig. \ref{fig:fail} shows examples where the model fails to geolocalize the video frames within 25km and predicts the model to be in a different city altogether. 


\end{document}